\documentclass[11pt]{article}

\usepackage[preprint]{acl}

\usepackage{times}
\usepackage{latexsym}

\usepackage[textsize=tiny]{todonotes}

\usepackage[T1]{fontenc}

\usepackage[utf8]{inputenc}

\usepackage{microtype}

\usepackage{inconsolata}
\usepackage{lipsum}

\usepackage{graphicx}
\usepackage{amsmath,amssymb}
\usepackage{booktabs}
\usepackage{tabularx}
\usepackage{array}
\usepackage{enumitem}
\usepackage{xcolor}
\usepackage{listings}
\usepackage{cleveref}
\usepackage{subcaption}
\usepackage[most]{tcolorbox}
\definecolor{promptbg}{HTML}{F5F6F8}
\definecolor{promptaccent}{HTML}{3F6BA8}
\newcommand{\plabel}[1]{\textcolor{promptaccent}{#1}}

\title{Why is AI-generated language so hard to understand? 
}
\author{Thomas Hikaru Clark$^{1}$ \quad \quad Matias B. Martin$^{2}$ \quad \quad Kyle Mahowald$^{2}$ \\
  $^{1}$New York University \quad \quad $^{2}$The University of Texas at Austin \\
  \texttt{thomas.hikaru.clark@nyu.edu \quad \{matias.martin,kyle\}@utexas.edu}}

\begin{document}
\maketitle
\begin{abstract}
There is a widespread sense that AI-generated text has recently become harder to understand.
In this study, we investigate a set of psycholinguistic features that are known to correlate with human processing difficulty (e.g., object relative clauses, long syntactic dependencies, low-frequency n-grams).  
Using a novel corpus of AI-generated text across diverse domains, we find that AI models use these difficult features more often than paraphrases generated by round-trip translation. Compared with human writing, all nine frontier and near-frontier models use more inanimate subjects with animate verbs and more low-frequency n-grams, and most also use longer dependencies, more stacked nouns, and more reduced object relatives in conversation; we also demonstrate substantial variation across model families.
We propose and test a hypothesis as to why: while human language is optimized for successful communication, AI-generated text is increasingly optimized for other goals like math, code, and reasoning, in which there is a pressure for \textit{information density} rather than successful communication.
We provide a proof-of-concept for this hypothesis by fine-tuning a language model under a length penalty, which leads to a divergence from human-like linguistic behavior.
Our findings highlight the axes of misalignment between linguistic processing demands in humans and LLMs, and point toward possible remedies.\looseness=-1
\end{abstract}

\begin{figure}[t!]
        \centering
    \includegraphics[width=0.95\linewidth]{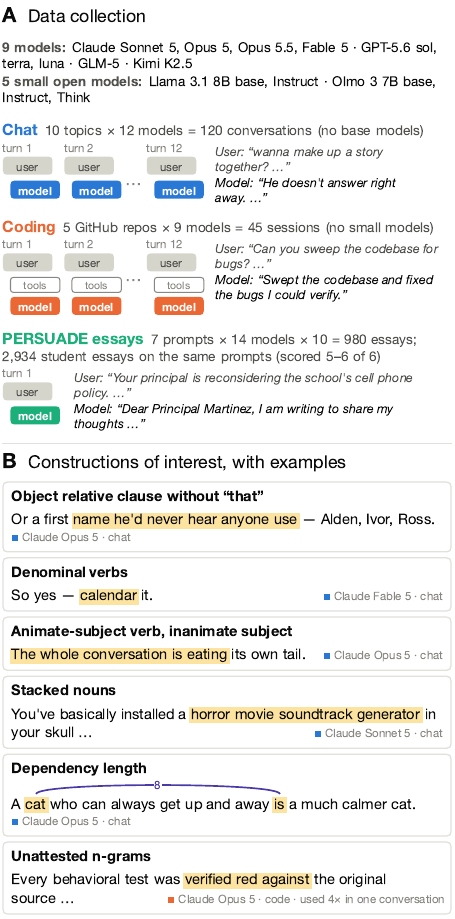}
    \caption{Overview of models studied, data collection procedure, and linguistic features under analysis.}
    \label{fig:overview}
\end{figure} 

\section{Introduction}

\emph{Your instinct is right; the convoluted sentence you are currently reading is quietly eating its own reader-patience budget --- footnote it.}
If you saw that sentence in the wild, you might suspect that it's the product of a modern LLM. 
It has several linguistic features that are overrepresented in frontier LLM text: an object relative clause, a denominal verb (``footnote''), an inanimate subject used with the verb ``eat'', and stacked nouns leading to a low-frequency n-gram sequence (``reader-patience budget'').
All of these features are associated with linguistic processing difficulty.\looseness=-1

But why would models trained on human text develop linguistic features that hurt intelligibility? And, if they are, is it happening systematically across model families?
Here, we study a set of features known to correlate with linguistic processing difficulty and find that they are overrepresented in text generated from recent LLMs, including both closed and open-weight models. 
We collect a novel corpus of LLM-generated text across diverse domains (chat, coding, essay), and we show that the prevalence of these features, in many models, exceeds that of human-generated reference corpora. 
Using a round-trip translation paradigm, we further show that these features do not occur in paraphrases that are controlled tightly for semantic content---further evidence they are a result of the specific features of modern post-trained models.\looseness=-1

Having established this trend, we then propose and test a hypothesis as to \textit{why} this pattern may have emerged in LLMs.
The psycholinguistics literature shows that human language is largely optimized for communication \citep{gibson2019how}, in the sense of striking an efficient balance between being easy to produce and easy to understand.
Empirically, human text is well-optimized relative to random baselines on a number of dimensions, such as minimizing dependency length \citep{futrell2015largescale}.
It follows that models trained to mimic human-generated text should also show a similar profile of linguistic features in their output, but that these patterns may not be preserved in models where communication is not the primary objective, and where humans are largely out of the loop.\looseness=-1

Indeed, modern LLMs are not just trained on human-generated text, but undergo extensive post-training for math, coding, and other reasoning tasks.
In paradigms like  those of \citet{deepseekai2025deepseekr1}, models are rewarded for getting the right answer but not necessarily for clarity.
We hypothesize that optimization for these goals, alongside a limit on the context window available to models, leads to information compression in a way that competes with intelligibility.
To test that hypothesis, we fine-tune a pre-trained language model under a length penalty, and measure the linguistic profile of its outputs. The model learns to succeed at the task using fewer tokens by leveraging syntactic alternations (cases where there are two ways of expressing the same meaning), even when this makes their linguistic output less humanlike. Furthermore, their preference for information compression bleeds into linguistic alternations that were not seen in training.\looseness=-1

In sum, this work tests two specific hypotheses about the intelligibility of AI-generated text. 
\textbf{H1:} The language output of frontier and near-frontier LLMs overuses linguistic features associated with human comprehension difficulty.
\textbf{H2:} A pressure for information density during post-training shifts the language output of LLMs to be less humanlike.
Using a combination of linguistic analysis and direct post-training intervention, we find support for both hypotheses. Our findings highlight the dimensions along which human language processing and frontier LLMs are misaligned, and point to possible interventions for addressing this misalignment.\looseness=-1

\section{Background and Related Work}

Recent scientific work has attempted to measure the idiosyncrasies of LLMs, such as their tendency to over-use particular words \citep{juzek2025why}, or their dense, noun-heavy style \cite{reinhart2025llms}. 
These idiosyncrasies may emerge during the post-training process of modern LLMs, including rewards during reinforcement learning from human feedback (RLHF; \citealp{ouyang2022training, singhal2024long}), or reinforcement learning with verifiable rewards (RLVR; \citealp{lambert2024tulu, deepseekai2025deepseekr1}).
The distinctive tells of LLM-generated writing have led to AI-generated text detectors \citep{bao2024fastdetectgpt}, and have raised questions regarding the impact of such AI-generated writing on linguistic content, with implications for diversity and representation \citep{padmakumar2024writing, agarwal2025ai, ryan2024unintended}. 
As a growing share of the text people read is shaped by LLMs, the linguistic properties of AI-generated writing matter not only for detection but also for the human readers who must process it. Yet whether these properties make AI-generated text harder for humans to understand has received little academic attention, despite widespread anecdotal reports of issues with LLM intelligibility \citep{uwaterloo2026claudish}.\looseness=-1

The field of psycholinguistics has extensively studied the linguistic features that drive comprehension difficulty. Words that are hard to predict are costly to process \citep{hale2001probabilistic, levy2008expectation}, and so are long-distance dependencies, because of limits on memory \citep{gibson1998linguistic, gibson2000dependency}. Memory-based accounts explain the difficulty of center-embedded sentences \citep{gibson1999memory, lewis1996interference} and of legalese \citep{martinez2022poor}, and temporary ambiguity can derail comprehension in garden-path sentences \citep{christianson2001thematic}. 
\looseness=-1

Crucially, the human mind operates under distinct pressures and limitations compared with those of LLMs. 
LLMs are also trained on significantly more data than humans experience during language development \cite {frank-2023-child-LLM}. Modern LLMs are able to attend fully to very large context windows \citep{vaswaniAttentionAllYou2017, ohModelHumanLinguistic2026}, in contrast with the lossy and distance-sensitive memory of humans \citep{futrell2020lossy}. Larger LLMs often fit human behavior (e.g., reading times)  less well than smaller LLMs \citep{oh2023why, oh2024frequency}. Consistent with this pattern, LLMs with artificially constrained memory fit human behavior better \citep{kuribayashi2022context,xu-etal-2026-memory,clarkLinearRecencyBias}. This suggests that LLMs' ability to attend fully to long contexts may come at the cost of human-likeness.\looseness=-1

Linguistic alternations, in which speakers and writers have multiple options for expressing a given meaning, provide a source of evidence for the preferences guiding human language generation.
One purported pressure on human language production is   uniform information density (UID) --- the idea that speakers spread information evenly across the signal to avoid overloading comprehenders, e.g. by inserting optional words \citep{ levy2007speakers, meister2021revisiting}. 
UID is at odds with \textit{maximizing} information density, by which shorter members of linguistic alternations would be categorically preferred, and has been proposed as a metric for preference optimization in LLM training to produce more human-aligned text \citep{liUniformInformationDensityBased2026}. 
Meanwhile, psycholinguistic evidence also supports a pressure based on cognitive availability during incremental production \citep{bockEffectAccessibilityWord1987, clarkEvidenceAvailabilityEffects2022, futrellInformationTheoreticAccountAvailability2024}.
Pre-trained LLMs inherit these signatures from their human training text, but post-training for specific tasks (e.g. math, code) may favor information-dense constructions over those that ease human processing or resemble humans.\looseness=-1

\section{Experiment 1: Linguistic features in AI-generated conversations}

\subsection{Linguistic features of interest}
\label{sec:features}

The purpose of Experiment 1 is to test \textbf{H1}: do frontier and near-frontier LLMs disproportionately use linguistic features that are linked to human processing difficulty? 
For each feature, we justify its association with processing difficulty, and explain why it might plausibly arise from a pressure toward information density. Example sentences for each feature are given in \Cref{tab:examples}. All features are identified from spaCy dependency parses \citep{honnibalSpaCyIndustrialstrengthNatural2020}; for denominal verbs and inanimate subjects, parser-generated candidates are then judged by an automated classifier (\Cref{sec:app-detection}).\looseness=-1

\paragraph{N-gram statistics} Predictable words are associated with lower processing difficulty, as measured by reading times \citep{levy2008expectation, smith2013effect}.
From an information-theoretic perspective, a word with a very high n-gram probability is more redundant than one with a low n-gram probability. 
For a system designed to maximize information density, we might expect high-probability n-grams to be reduced (i.e. common, uninformative, and predictable words being skipped), and higher overall n-gram perplexity (when measured by an independent language model).
Against a reference corpus, we measure the rate of \emph{unattested pairs}, bigrams that never occur in it, and \emph{perplexity} under a trigram model (\Cref{sec:app-baselines}).\looseness=-1

\paragraph{Dependency length} Words in a sentence relate to each other via syntactic dependency relations (e.g., between a noun and any determiners or adjectives that modify it, or between a verb and its subject, object, and adverbial modifiers). 
Long-distance dependencies are known to be difficult to process for humans, due to the additional strain they place on working memory \citep{gibson1998linguistic, gibson2000dependency, grodner2005consequences}, and speakers tend to avoid them \citep{futrell2015largescale, temperley2018minimizing}.
We standardize dependency length across sentence lengths, as described in \Cref{sec:app-analysis-details}.\looseness=-1

\paragraph{Object relative clauses} 
We measure the rate of object relative clauses (ORCs) in the language output of LLMs, differentiating between ORCs with an overt relativizer (most often \emph{that} or \emph{which}) and without one.
ORCs are known to present additional processing difficulty compared to subject relative clauses, potentially due to their non-canonical word order and longer dependencies \citep{staub2010eye}, although animacy and frequency reduce the gap \citep{mak2006animacy, reali2007processing}.
Under the hypothesis that LLMs tend to choose more compressed realizations of a given meaning, we predict object relative clauses will be over-represented, because they tend to replace a (longer) passive construction (\emph{The house that the contractors built} vs. \emph{The house that \textbf{was} built \textbf{by} the contractors}), and second, that \textit{reduced} relative clauses (without the optional relativizer) would be over-represented, because they avoid an optional word.\looseness=-1

\begin{figure}[t]
        \centering
    \includegraphics[width=\linewidth]{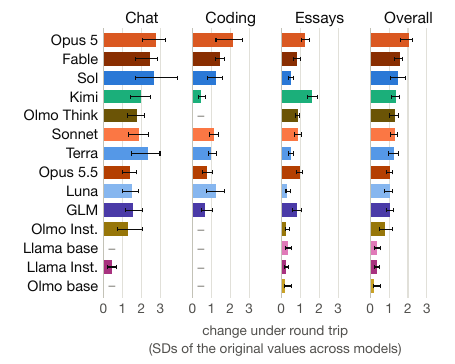}
\caption{Change in linguistic features under round-trip translation (NLLB-200 via Japanese), by model, in standard deviations of the original values across models (\Cref{sec:app-figure-details}). Bars: 95\% bootstrap intervals; a dash marks a setting in which the model was not run.}
\label{fig:roundtrip-change}
\end{figure}

\paragraph{Stacked noun compounds} 
In English, nouns freely modify other nouns in an adjectival way: in the phrase \emph{coffee shop}, \emph{coffee} modifies \emph{shop}, despite usually appearing as a noun. 
This feature enables stacking multiple nouns into rich compound words which denote highly specific meanings. 
However, such stacks can introduce bracketing ambiguity, as in the headline \textit{Squad Helps Dog Bite Victim} \citep{cooper1980squad}, and disrupt reading \citep{lauer-1995-corpus, nakov-hearst-2005-search, staubTimeCoursePlausibility2007, cohenOnlineProcessingNovel2014}.
Here, we measure the rate of occurrence of stacked noun compounds, defined as three or more nouns in a modifier relationship.\looseness=-1 

\paragraph{Denominal verbs} Verbs which are formed directly from nouns, such as in the phrase \emph{\textbf{bottle} the whiskey} are known as \emph{denominal} verbs.
While some denominal verbs are highly conventionalized and well-attested in human language, using a denominal verb in an ad-hoc way is marked (e.g., \textit{calendar} it). Readers can interpret such innovations \citep{clark1979nouns}, but part-of-speech conversion triggers syntactic re-evaluation \citep{thierry2008shakespearean}, and noun/verb-ambiguous words are processed differently from unambiguous ones \citep{frazier1987resolution, federmeier2000brain}.
Often, the alternative to using a denominal verb is to use a longer, more syntactically complex phrase, e.g. \emph{put the whiskey in a bottle}.  
We therefore predict that a pressure for information compression will favor usage of denominal verbs beyond their base rate in humans, even at the expense of intelligibility.\looseness=-1

\looseness=-1 
\begin{figure*}
    \centering
    \begin{subfigure}[b]{0.56\linewidth}
        \centering
        \includegraphics[width=\linewidth]{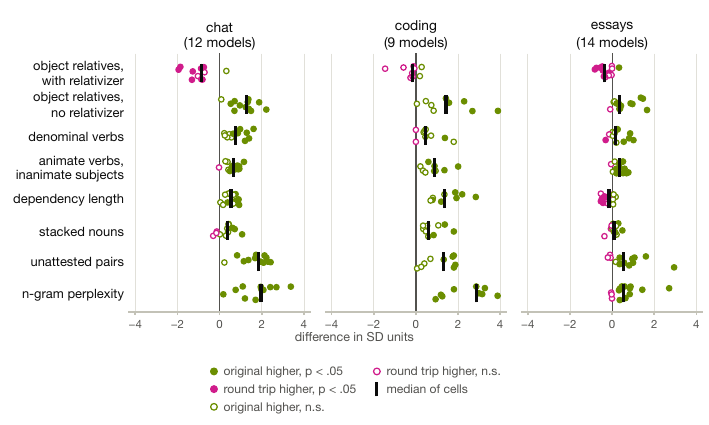}
        \caption{By setting: chat, coding, and essay.}
        \label{fig:feature-consistency-settings}
    \end{subfigure}
    \hfill
    \begin{subfigure}[b]{0.40\linewidth}
        \centering
        \includegraphics[width=.9\linewidth]{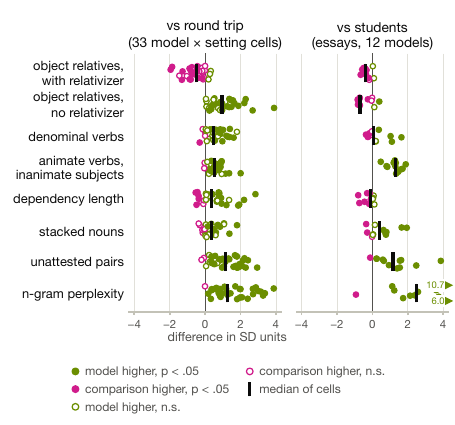}
        \caption{Round-trip translation vs.\ student essays.}
        \label{fig:feature-consistency-roundtrip-v-human}
    \end{subfigure}
    \caption{Summary of changes to each linguistic feature across applicable models. (a) Changes across the chat, coding, and essay settings. (b) Comparison of round-trip translation vs.\ student essays as the reference.}
    \label{fig:feature-consistency-combined}
\end{figure*}

\paragraph{Inanimate subjects} We count the instances of inanimate subjects used with active verbs that typically take only animate subjects. Examples include ``\verb|is_file()| \textbf{swallows} that and returns False'' and ``Puzzle patterns where this asymmetry \textbf{bites}''. 
Such sentences violate a selectional restriction \citep{chomsky1965aspects} on which verbs typically take which subjects. They can be understood as personification or metonymy \citep{lakoff1980metaphors}, but readers use animacy as a cue during parsing \citep{trueswell1994semantic}, and without additional context, a mismatch may elicit a P600 response associated with reanalysis \citep{kim2005independence, kuperberg2007neural}.
However, rephrasing these sentences to avoid inanimate subjects would potentially require lengthening the sentence: for example, the word ``bites'' stands in for ``causes an issue''. 
We speculate that during training, models may have converged to using such constructions as efficient, albeit non-standard, ways of expressing meaning which would otherwise take up more tokens.\looseness=-1

\subsection{Data collection}

We collect a corpus of AI-generated conversations (\Cref{fig:overview}) using a set of 14 frontier and non-frontier LLMs: Kimi K2.5, GLM 5, Claude (Sonnet 5, Opus 5, Opus 5.5, Fable 5), and GPT-5.6 (Luna, Terra, Sol), Olmo \citep{olmo2025olmo3} (Base, Instruct, Think), and Llama \citep{grattafiori2024llama} (Base, Instruct).
We use a total of 15 conversational starting prompts, or \textit{seeds}, consisting of 10 chat seeds and 5 coding seeds. Each coding seed involves a request to clone a real, public GitHub repo, identify bugs, and implement fixes within a sandbox environment. Chat sessions were capped at 12 turns (a simulated human query and a model response) and coding sessions at 11 turns plus a fixed closing message. In a third setting, each model writes single-turn essays to seven PERSUADE 2.0 assignments. Full data collection procedures and API costs are described in \Cref{sec:app-data-collection}.\looseness=-1

\subsection{Comparison to human baselines}
Where available, we compute a human baseline for the linguistic feature of interest, on approximately genre-matched text: OpenSubtitles \citep{lison-tiedemann-2016-opensubtitles} for chat, StackExchange for coding, and student essays from the PERSUADE 2.0 corpus \citep{crossley2024persuade} for essays. We additionally include a Wikipedia comparison.
The same corpora serve as the references for the n-gram measures (\Cref{sec:app-baselines}).\looseness=-1

\subsection{Round-trip translation}

We employ round-trip translation (bilingual pivoting), an established paraphrasing method that preserves meaning at scale \cite{bannard-callison-burch-2005-paraphrasing, mallinson-etal-2017-paraphrasing, wieting-gimpel-2018-paranmt, federmann-etal-2019-multilingual}, as a method for separating a model output's characteristic syntactic profile from its semantics. Translation normalizes the output to frequent forms, reducing lexical and syntactical richness \cite{vanmassenhove-etal-2019-lost, vanmassenhove-etal-2021-machine} and stripping stylistic and authorial traits \cite{rabinovich-etal-2017-personalized, prabhumoye-etal-2018-style}. 
The result is a parallel corpus in which each translation is paired with its source sentence. Because the round trip holds the semantic content of each response fixed while replacing its wording, a feature that becomes less frequent after the round trip is one that the model chose over an available alternative way of expressing the same content. 
To create this parallel corpus, we translate each sentence independently from English to Japanese and back with NLLB-200 (3.3B; \citealp{nllbteam2024scaling}). 
Japanese was chosen because pivot languages unrelated to English tend to yield more diverse paraphrases \cite{federmann-etal-2019-multilingual}. 

\begin{figure*}
    \centering
    \includegraphics[width=0.9\linewidth]{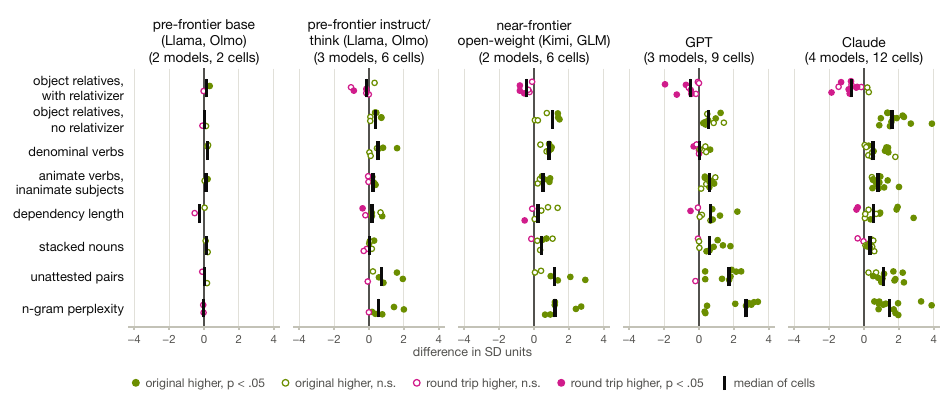}
    \caption{Summary of changes to each linguistic feature across applicable models across the following model families: base pre-frontier models, post-trained pre-frontier models, near-frontier open-weight models, GPT, and Claude. Each ``cell'' denotes a combination of model and setting (e.g. chat, coding, or essay). Not all models are compatible with all settings (e.g. base models only produced essays, not chat or coding results). Model preferences for objective relatives with and without a relativizer tend to point in opposite directions, suggesting a consistent preference for the no-relativizer alternative.}
    \label{fig:feature-consistency-category}
\end{figure*}
\subsection{Results: AI models overuse constructions with known processing costs}

\begin{figure*}
        \centering
    \includegraphics[width=.98\linewidth]{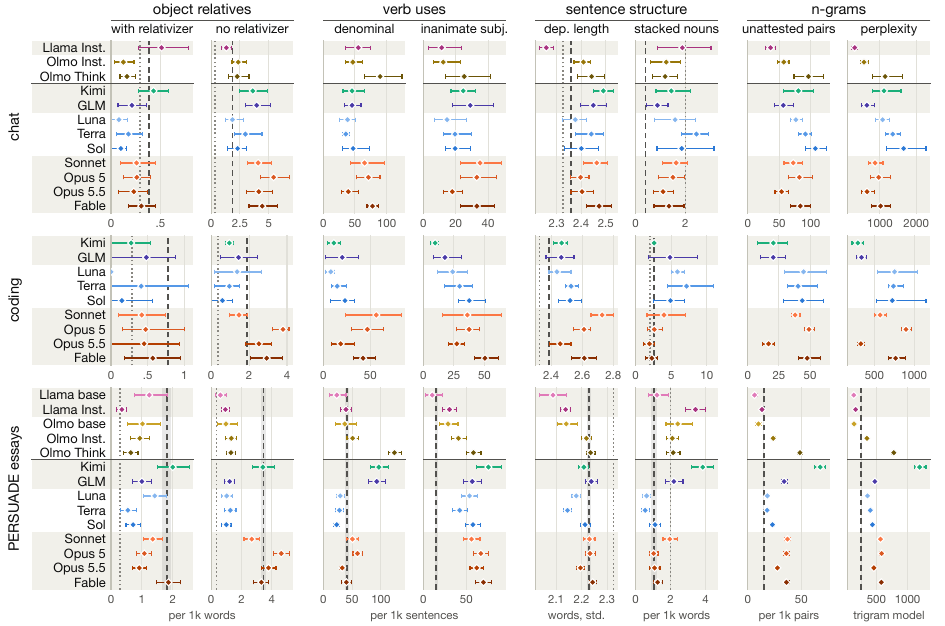}
    \caption{%
  \textbf{Comparison of linguistic features across models and settings.} Diamonds: each model's pooled value; bars: 95\% bootstrap intervals. Dashed lines: human baseline from OpenSubtitles (chat), StackExchange (coding), and PERSUADE 2.0 student essays (essays). Dotted lines: Wikipedia. Details and panels without a human baseline: \Cref{sec:app-figure-details}. Summary figures for by-model and by-feature comparison are provided in \Cref{fig:roundtrip-change,fig:feature-consistency-category,fig:feature-consistency-combined}
 }
\label{fig:combined-figure}
\end{figure*}

\begin{figure*}
    \centering
    \includegraphics[width=.95\linewidth]{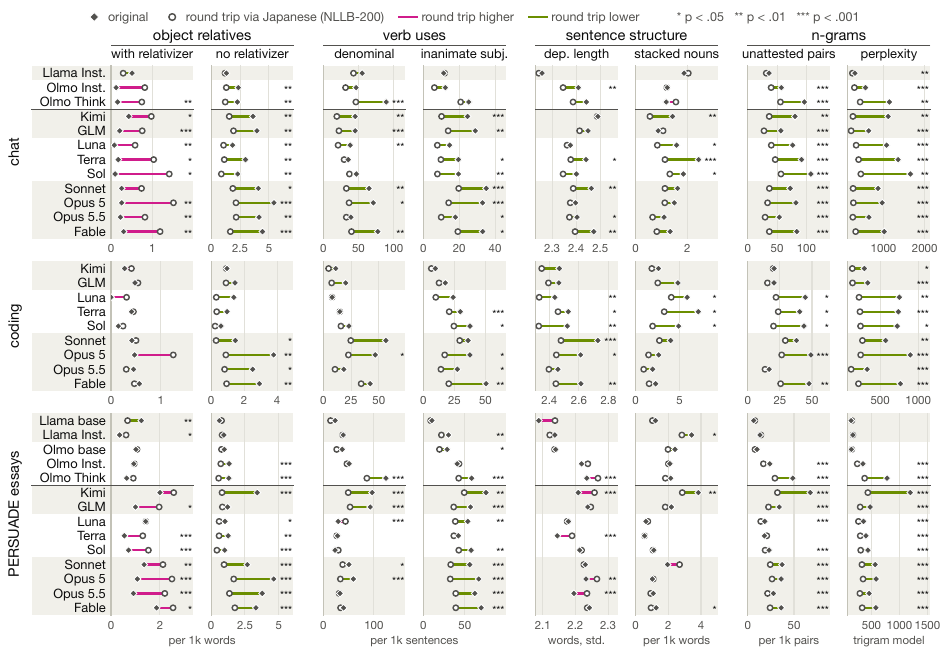}
    \caption{Comparison of linguistic features across models and settings, relative to a parallel corpus generated with round-trip translation via Japanese (NLLB-200). Stars: paired test of the difference (\Cref{sec:app-figure-details}). 
    For most features and most models, responses over-use the feature relative to the round-trip translation. 
    \looseness=-1}
    \label{fig:round-trip}
\end{figure*}

\Cref{fig:combined-figure} compares the linguistic features of interest across the three settings of Chat, Code, and Essay, and 14 LLMs (12 in chat, 9 in coding, and all 14 in essays).
\Cref{fig:round-trip} compares the LLM-generated corpus to the parallel corpus derived via round-trip translation. 
We highlight several results.\looseness=-1

\paragraph{Frontier models over-use rare n-grams, denominal verbs, inanimate subjects, and reduced ORCs, relative to the paraphrase distribution.}
Across all three settings and almost all frontier and near-frontier models, the round-trip translations reveal that AI text has more unattested bigrams and higher n-gram perplexity. These effects are absent or small for the Llama models and the Olmo base model, but Olmo Think shows effects as large as those of the frontier models. Within the Olmo family, the effects grow from the base to the instruct to the reasoning model.
The same pattern largely holds for denominal verbs and inanimate subjects, as well as for ORCs with no relativizer. Meanwhile, models under-use ORCs \emph{with} the relativizer, suggesting a preference for skipping optional words (\Cref{fig:feature-consistency-category}).
Similar patterns hold for stacked nouns and dependency length in chat and coding. In the essays, dependency length is unchanged or slightly longer after the round trip, and stacked nouns change significantly for only two models.\looseness=-1

\paragraph{These effects are smaller overall for the PERSUADE essays.} The PERSUADE essays show smaller changes under round-trip translation on several categories (n-gram measures, dependency lengths, stacked nouns, denominal verbs) relative to chat and coding (\Cref{fig:round-trip,fig:feature-consistency-settings}). We hypothesize that this is because they involve a single prompt, unlike the many iterated turns of chat and coding. We speculate that certain patterns may emerge in long back-and-forth conversations (which may demand more information compression).\looseness=-1

\paragraph{There are notable differences between models.}
The Llama models and the Olmo base and instruct models change least under round-trip translation (\Cref{fig:roundtrip-change}) and are closest to the student essays on the n-gram and verb measures. Meanwhile, Olmo Think patterns with the frontier models.
Notably, Claude Opus 5.5 is lower than its predecessor Opus 5 on nearly every difficulty-linked measure in every setting (19 of 21 comparisons), with non-overlapping 95\% intervals in eight.
In sum, most frontier models over-represent linguistically difficult features relative to paraphrases, and relative to human text for inanimate subjects and low-frequency n-grams (\Cref{fig:feature-consistency-roundtrip-v-human}).\looseness=-1

\begin{figure*}
        \centering
    \includegraphics[width=0.9\linewidth]{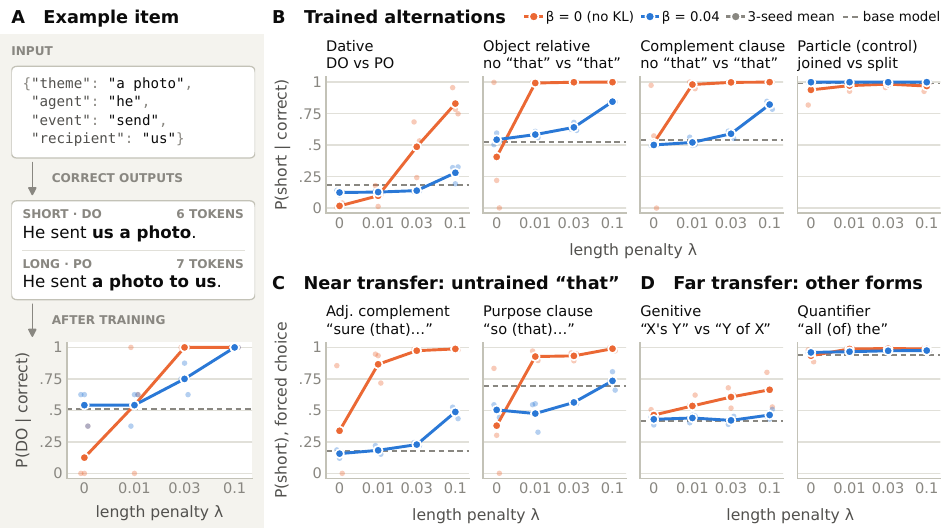}
\caption{\textbf{Effect of length penalty during RL on linguistic preferences.} (A) Example item. (B) Share of correct sampled outputs using the short variant, on unseen items. (C, D) Forced-choice probability of the short variant. Points: mean of three seeds (faint: single seeds); dashed: base model; orange: $\beta = 0$ (no KL); blue: $\beta = 0.04$.}
\label{fig:length-penalty}
\end{figure*}

\section{Experiment 2: Effect of length penalties on language preferences}

The purpose of Experiment 2 is to test \textbf{H2} -- the hypothesis that models trained under a pressure for information density are more likely to shift their linguistic output toward difficult and non-humanlike linguistic features. While Experiment 1 provides circumstantial evidence for \textbf{H2} (e.g., that models over-use constructions that are less verbose than semantically equivalent alternatives), we additionally aim to understand a possible mechanism by which a pressure for information density may arise in LLMs.\looseness=-1

\subsection{Setup}
We start with an open-weights model, \verb|Qwen2.5-1.5B-Instruct| \citep{qwen2024qwen25}, and a task of converting JSON-style structured event descriptions into English sentences. 
We apply fine-tuning via reinforcement learning with GRPO \citep{shao2024deepseekmath}, varying two key parameters: $\beta$, the KL divergence regularization term which penalizes a model for drifting from the base model's next-token distribution, and $\lambda$, a length penalty term which penalizes responses for verbosity. 
Each event seen during training or testing can be expressed using either member of a syntactic alternation where one alternative is shorter by a known number of tokens (except the particle control, where both orders have the same length). 
Human speakers use both members of such alternations and choose between them according to context; for example, they are more likely to include an optional \emph{that} when the material that follows is less predictable \citep{levy2007speakers, jaeger2010redundancy}.
The key question is whether training a model with a length penalty pushes it instead toward a categorical preference for the shorter member, regardless of context, which would be a divergence from human language use.
We evaluate in three settings. The \textbf{trained alternations} are new sentences, never seen in training, built from the four alternations used in training. \textbf{Near transfer} denotes test sentences which contain alternations involving the optional word ``that'' (\textit{sure (that)}, \textit{so (that)}); however, these specific constructions were not seen during training, so the model omitting ``that'' in these cases would indicate a generalization to a novel context. Finally, \textbf{far transfer} denotes unrelated alternations (genitive, quantifier, passive and purpose-clause alternations; \Cref{tab:alts}), where a model could in theory choose to conserve tokens by choosing the shorter alternation, though this would indicate that the model has generalized to an information-dense preference in contexts unrelated to the training data. See \Cref{sec:app-rl-details} for details (\Cref{fig:prompt}; \Cref{tab:hparams}).\looseness=-1

\subsection{Results: A length penalty shifts models toward shorter variants, including in untrained constructions}

As can be seen in Figure \ref{fig:length-penalty}, the probability of using the shorter construction increases as a function of the length penalty $\lambda$, both with and without the KL penalty $\beta$.
For the setting without KL-regularization, preference for the short alternative rises rapidly as a function of length penalty ($\lambda$). 
Preference for the short alternation rises, albeit more gradually, even in the presence of KL-regularization.
Without KL regularization, runs with $\lambda = 0$ had different preferences for the short alternative ranging from 0\% to 100\% depending on seed, indicating no stable preference. Meanwhile, for any $\lambda > 0$, all seeds converged to a categorical preference for the short alternative.
The models learn a tendency for information compression that transfers to additional constructions not seen in training. This was consistent in the near transfer case; far transfer was only observed without KL penalty in the genitive alternation, one of four far-transfer alternations tested (\Cref{sec:app-rl-details}).\looseness=-1

\subsection{Discussion}
We take these results as a proof of concept that a length penalty used during reinforcement learning with GRPO can directly affect grammatical features.
We note that the model was not directly instructed to use compressed or dense language, and was not directly given information about the availability of syntactic alternations. Rather, the penalty resulted in an emergent preference for shorter members of syntactic alternations. Without KL regularization this preference was near-categorical (over 98\% short for both trained ``that'' alternations at every $\lambda > 0$), unlike human speakers, who choose between the variants by context. Crucially, the models were still able to succeed at their task, since both members of an alternation were valid solutions to the training task. We speculate that the use of verifiable rewards creates an environment where correctness is decoupled from humanlike communication, leading models to converge toward a novel, potentially less comprehensible distribution of language use.\looseness=-1

\section{General Discussion}

\paragraph{Model Development} Past work has identified RLHF as a potential factor in the rise of word overuse by AI \citep{juzek2025why, juzek2026word}.
At the same time, a movement from RLHF to RLVR removes humans from the loop, and rewards models for verifiable solutions to problems. Human-feedback training has in fact been found to reward longer outputs \citep{singhal2024long}, so the pressure toward compression that we describe is more plausibly tied to verifiable-reward training. The move away from human feedback removes the implicit pressure for successful communication, and we speculate that this allows LLM-generated language to converge toward a dialect that is useful for LLMs to compress information densely and solve complex problems, but is suboptimal from the perspective of human intelligibility. 
It is of course also possible to reward models for effective communication; indeed, our results indicate that more recent model variants like Opus 5.5 yield a quantifiable improvement over older models like Opus 5.\looseness=-1

\paragraph{Implications for psycholinguistics}
Multiple threads of research have studied the pressures that have shaped the structure of languages' grammars and lexicons over the course of language evolution. Various pressures have been identified, such as those for dependency length minimization \citep{gildea2010grammars, temperley2018minimizing}, communicative efficiency \citep{hahnUniversalsWordOrder2020}, and uniform information density \citep{clarkCrossLinguisticPressureUniform2023}. 
These pressures are not necessarily absolute (i.e. holding in all contexts in all languages), but are supported by large-scale, aggregate evidence across many of the world's languages. 
A key takeaway of this literature is that actual human language use is constrained by the realities and limitations of human comprehension, such as lossy working memory or incremental processing; sentences which consistently fail to lead to communicative success are likely to be weeded out from a language via grammar or convention. 
The wide-scale training and deployment of AI models in contexts where human comprehension is not the primary purpose -- such as developing large code bases or solving mathematical problems -- presents a fundamentally different set of evolutionary pressures upon the linguistic profile of AI-generated language compared to natural language, and potentially provides novel opportunities for studying the emergence of communication systems. \looseness=-1

\section{Conclusion}

For frontier LLMs, the training objectives that make them successful at tasks like agentic coding and mathematical reasoning, especially with long and expensive context windows, are not guaranteed to be aligned with the same pressures that influence human language. 
Natural language is subject to a pressure for communicative success, assuming the cognitive resource constraints of the human mind and incremental processing. 
This discrepancy may help explain why many frontier LLMs use linguistic constructions which are technically allowed by the grammar of English, yet are out-of-distribution for human language and are associated with comprehension difficulty. 
While increased exposure to LLM-generated text may acclimate humans to these out-of-distribution constructions, many of the patterns that have been evaluated in this work present fundamental challenges to intelligibility, and future work on LLMs may consider methods for better aligning LLM outputs with the needs of human comprehenders, in the service of user experience and monitorability.\looseness=-1

\section{Limitations}

We only examine a subset of commonly used models, only conduct analyses in English, and use a limited number of conversation seeds (chat topics and code repositories). Ideally, the study would be replicated on a larger number of seeds and model families.
Causally identifying sources of linguistic difficulty in frontier models would ideally involve studying them as they are post-trained, but this is not feasible to do at scale and so we rely on proxies. Similarly, limited compute makes studying post-training difficult.

\section{Acknowledgments}

K.M. acknowledges support from NSF CAREER grant 2339729. We thank Sasha Boguraev, Kanishka Misra, Elias Stengel-Eskin, and Qing Yao for helpful discussions.

\bibliography{custom}

\begin{thebibliography}{82}
\providecommand{\natexlab}[1]{#1}

\bibitem[{Agarwal et~al.(2025)Agarwal, Naaman, and Vashistha}]{agarwal2025ai}
Dhruv Agarwal, Mor Naaman, and Aditya Vashistha. 2025.
\newblock \href {https://doi.org/10.1145/3706598.3713564} {{AI} suggestions
  homogenize writing toward {W}estern styles and diminish cultural nuances}.
\newblock In \emph{Proceedings of the 2025 {CHI} Conference on Human Factors in
  Computing Systems}, pages 1--21.

\bibitem[{Bannard and
  Callison-Burch(2005)}]{bannard-callison-burch-2005-paraphrasing}
Colin Bannard and Chris Callison-Burch. 2005.
\newblock \href {https://doi.org/10.3115/1219840.1219914} {Paraphrasing with
  bilingual parallel corpora}.
\newblock In \emph{Proceedings of the 43rd Annual Meeting of the Association
  for Computational Linguistics ({ACL}{'}05)}, pages 597--604, Ann Arbor,
  Michigan. Association for Computational Linguistics.

\bibitem[{Bao et~al.(2024)Bao, Zhao, Teng, Yang, and
  Zhang}]{bao2024fastdetectgpt}
Guangsheng Bao, Yanbin Zhao, Zhiyang Teng, Linyi Yang, and Yue Zhang. 2024.
\newblock \href {https://arxiv.org/abs/2310.05130} {Fast-{DetectGPT}: Efficient
  zero-shot detection of machine-generated text via conditional probability
  curvature}.
\newblock In \emph{International Conference on Learning Representations}.

\bibitem[{Bock(1987)}]{bockEffectAccessibilityWord1987}
Kathryn Bock. 1987.
\newblock \href {https://doi.org/10.1016/0749-596X(87)90120-3} {An effect of
  the accessibility of word forms on sentence structures}.
\newblock \emph{Journal of Memory and Language}, 26(2):119--137.

\bibitem[{Chomsky(1965)}]{chomsky1965aspects}
Noam Chomsky. 1965.
\newblock \emph{Aspects of the Theory of Syntax}.
\newblock MIT Press, Cambridge, MA.

\bibitem[{Christianson et~al.(2001)Christianson, Hollingworth, Halliwell, and
  Ferreira}]{christianson2001thematic}
Kiel Christianson, Andrew Hollingworth, John~F. Halliwell, and Fernanda
  Ferreira. 2001.
\newblock \href {https://doi.org/10.1006/cogp.2001.0752} {Thematic roles
  assigned along the garden path linger}.
\newblock \emph{Cognitive Psychology}, 42(4):368--407.

\bibitem[{Clark et~al.(2025)Clark, Oh, and Schuler}]{clarkLinearRecencyBias}
Christian Clark, Byung-Doh Oh, and William Schuler. 2025.
\newblock \href {https://aclanthology.org/2025.coling-main.517/} {Linear
  recency bias during training improves {Transformers}' fit to reading times}.
\newblock In \emph{Proceedings of the 31st International Conference on
  Computational Linguistics}, pages 7735--7747, Abu Dhabi, UAE. Association for
  Computational Linguistics.

\bibitem[{Clark and Clark(1979)}]{clark1979nouns}
Eve~V. Clark and Herbert~H. Clark. 1979.
\newblock \href {https://doi.org/10.2307/412745} {When nouns surface as verbs}.
\newblock \emph{Language}, 55(4):767--811.

\bibitem[{Clark et~al.(2022)Clark, Wilcox, Gibson, and
  Levy}]{clarkEvidenceAvailabilityEffects2022}
Thomas Clark, Ethan~Gotlieb Wilcox, Edward Gibson, and Roger Levy. 2022.
\newblock \href {https://escholarship.org/uc/item/1q19f8vt} {Evidence for
  {Availability} {Effects} on {Speaker} {Choice} in the {Russian} {Comparative}
  {Alternation}}.
\newblock In \emph{Proceedings of the {Annual} {Meeting} of the {Cognitive}
  {Science} {Society}}, volume~44.

\bibitem[{Clark et~al.(2023)Clark, Meister, Pimentel, Hahn, Cotterell, Futrell,
  and Levy}]{clarkCrossLinguisticPressureUniform2023}
Thomas~Hikaru Clark, Clara Meister, Tiago Pimentel, Michael Hahn, Ryan
  Cotterell, Richard Futrell, and Roger Levy. 2023.
\newblock \href {https://doi.org/10.1162/tacl_a_00589} {A cross-linguistic
  pressure for uniform information density in word order}.
\newblock \emph{Transactions of the Association for Computational Linguistics},
  11:1048--1065.

\bibitem[{Cohen and Staub(2014)}]{cohenOnlineProcessingNovel2014}
Andrew~L. Cohen and Adrian Staub. 2014.
\newblock \href {https://doi.org/10.1080/17470218.2013.796398} {Online
  processing of novel noun-noun compounds: eye movement evidence}.
\newblock \emph{Quarterly Journal of Experimental Psychology}, 67(1):147--165.

\bibitem[{Cooper(1980)}]{cooper1980squad}
Gloria Cooper, editor. 1980.
\newblock \emph{Squad Helps Dog Bite Victim, and Other Flubs from the Nation's
  Press}.
\newblock Dolphin Books/Doubleday, Garden City, NY.
\newblock Compiled from the \emph{Columbia Journalism Review}.

\bibitem[{Crossley et~al.(2024)Crossley, Tian, Baffour, Franklin, Benner, and
  Boser}]{crossley2024persuade}
Scott~A. Crossley, Yu~Tian, Perpetual Baffour, Alex Franklin, Meg Benner, and
  Ulrich Boser. 2024.
\newblock \href {https://doi.org/10.1016/j.asw.2024.100865} {A large-scale
  corpus for assessing written argumentation: Persuade 2.0}.
\newblock \emph{Assessing Writing}, 61:100865.

\bibitem[{Federmann et~al.(2019)Federmann, Elachqar, and
  Quirk}]{federmann-etal-2019-multilingual}
Christian Federmann, Oussama Elachqar, and Chris Quirk. 2019.
\newblock \href {https://doi.org/10.18653/v1/D19-5503} {Multilingual whispers:
  Generating paraphrases with translation}.
\newblock In \emph{Proceedings of the 5th Workshop on Noisy User-generated Text
  (W-NUT 2019)}, pages 17--26. Association for Computational Linguistics.

\bibitem[{Federmeier et~al.(2000)Federmeier, Segal, Lombrozo, and
  Kutas}]{federmeier2000brain}
Kara~D. Federmeier, Jessica~B. Segal, Tania Lombrozo, and Marta Kutas. 2000.
\newblock \href {https://doi.org/10.1093/brain/123.12.2552} {Brain responses to
  nouns, verbs and class-ambiguous words in context}.
\newblock \emph{Brain}, 123(12):2552--2566.

\bibitem[{Finkelstein et~al.(2026)}]{finkelstein2026translategemma}
Mara Finkelstein et~al. 2026.
\newblock \href {https://arxiv.org/abs/2601.09012} {{TranslateGemma} technical
  report}.
\newblock \emph{Preprint}, arXiv:2601.09012.

\bibitem[{Frank(2023)}]{frank-2023-child-LLM}
Michael~C. Frank. 2023.
\newblock \href {https://doi.org/10.1016/j.tics.2023.08.007} {Bridging the data
  gap between children and large language models}.
\newblock \emph{Trends in Cognitive Sciences}, 27(11):990--992.

\bibitem[{Frazier and Rayner(1987)}]{frazier1987resolution}
Lyn Frazier and Keith Rayner. 1987.
\newblock \href {https://doi.org/10.1016/0749-596X(87)90137-9} {Resolution of
  syntactic category ambiguities: Eye movements in parsing lexically ambiguous
  sentences}.
\newblock \emph{Journal of Memory and Language}, 26(5):505--526.

\bibitem[{Futrell(2024)}]{futrellInformationTheoreticAccountAvailability2024}
Richard Futrell. 2024.
\newblock \href {https://doi.org/10.1111/tops.12716} {An
  {Information}-{Theoretic} {Account} of {Availability} {Effects} in {Language}
  {Production}}.
\newblock \emph{Topics in Cognitive Science}, 16(1):38--53.

\bibitem[{Futrell et~al.(2020)Futrell, Gibson, and Levy}]{futrell2020lossy}
Richard Futrell, Edward Gibson, and Roger~P. Levy. 2020.
\newblock \href {https://doi.org/10.1111/cogs.12814} {Lossy-context surprisal:
  An information-theoretic model of memory effects in sentence processing}.
\newblock \emph{Cognitive Science}, 44(3):e12814.

\bibitem[{Futrell et~al.(2015)Futrell, Mahowald, and
  Gibson}]{futrell2015largescale}
Richard Futrell, Kyle Mahowald, and Edward Gibson. 2015.
\newblock \href {https://doi.org/10.1073/pnas.1502134112} {Large-scale evidence
  of dependency length minimization in 37 languages}.
\newblock \emph{Proceedings of the National Academy of Sciences},
  112(33):10336--10341.

\bibitem[{Gibson(1998)}]{gibson1998linguistic}
Edward Gibson. 1998.
\newblock \href {https://doi.org/10.1016/s0010-0277(98)00034-1} {Linguistic
  complexity: Locality of syntactic dependencies}.
\newblock \emph{Cognition}, 68(1):1--76.

\bibitem[{Gibson(2000)}]{gibson2000dependency}
Edward Gibson. 2000.
\newblock \href {https://doi.org/10.7551/mitpress/3654.003.0008} {The
  dependency locality theory: A distance-based theory of linguistic
  complexity}.
\newblock In Alec Marantz, Yasushi Miyashita, and Wayne O'Neil, editors,
  \emph{Image, Language, Brain: Papers from the First Mind Articulation Project
  Symposium}, pages 95--126. MIT Press, Cambridge, MA.

\bibitem[{Gibson et~al.(2019)Gibson, Futrell, Piantadosi, Dautriche, Mahowald,
  Bergen, and Levy}]{gibson2019how}
Edward Gibson, Richard Futrell, Steven~T. Piantadosi, Isabelle Dautriche, Kyle
  Mahowald, Leon Bergen, and Roger Levy. 2019.
\newblock \href {https://doi.org/10.1016/j.tics.2019.02.003} {How efficiency
  shapes human language}.
\newblock \emph{Trends in Cognitive Sciences}, 23(5):389--407.

\bibitem[{Gibson and Thomas(1999)}]{gibson1999memory}
Edward Gibson and James Thomas. 1999.
\newblock \href {https://doi.org/10.1080/016909699386293} {Memory limitations
  and structural forgetting: The perception of complex ungrammatical sentences
  as grammatical}.
\newblock \emph{Language and Cognitive Processes}, 14(3):225--248.

\bibitem[{Gildea and Temperley(2010)}]{gildea2010grammars}
Daniel Gildea and David Temperley. 2010.
\newblock \href {https://doi.org/10.1111/j.1551-6709.2009.01073.x} {Do grammars
  minimize dependency length?}
\newblock \emph{Cognitive Science}, 34(2):286--310.

\bibitem[{Grattafiori et~al.(2024)}]{grattafiori2024llama}
Aaron Grattafiori et~al. 2024.
\newblock \href {https://arxiv.org/abs/2407.21783} {The {Llama} 3 herd of
  models}.
\newblock \emph{Preprint}, arXiv:2407.21783.

\bibitem[{Grodner and Gibson(2005)}]{grodner2005consequences}
Daniel Grodner and Edward Gibson. 2005.
\newblock \href {https://doi.org/10.1207/s15516709cog0000_7} {Consequences of
  the serial nature of linguistic input for sentenial complexity}.
\newblock \emph{Cognitive Science}, 29(2):261--290.

\bibitem[{Guo et~al.(2025)Guo, Yang, Zhang, Song, Wang, Zhu, Xu, Zhang, Ma, Bi
  et~al.}]{deepseekai2025deepseekr1}
Daya Guo, Dejian Yang, Haowei Zhang, Junxiao Song, Peiyi Wang, Qihao Zhu,
  Runxin Xu, Ruoyu Zhang, Shirong Ma, Xiao Bi, et~al. 2025.
\newblock \href {https://doi.org/10.1038/s41586-025-09422-z} {{DeepSeek-R1}
  incentivizes reasoning in {LLMs} through reinforcement learning}.
\newblock \emph{Nature}, 645(8081):633--638.
\newblock ArXiv preprint title: ``DeepSeek-R1: Incentivizing Reasoning
  Capability in LLMs via Reinforcement Learning''.

\bibitem[{Hahn et~al.(2020)Hahn, Jurafsky, and
  Futrell}]{hahnUniversalsWordOrder2020}
Michael Hahn, Dan Jurafsky, and Richard Futrell. 2020.
\newblock \href {https://doi.org/10.1073/pnas.1910923117} {Universals of word
  order reflect optimization of grammars for efficient communication}.
\newblock \emph{Proceedings of the National Academy of Sciences},
  117(5):2347--2353.

\bibitem[{Hale(2001)}]{hale2001probabilistic}
John Hale. 2001.
\newblock \href {https://doi.org/10.3115/1073336.1073357} {A probabilistic
  {E}arley parser as a psycholinguistic model}.
\newblock In \emph{Proceedings of the Second Meeting of the North American
  Chapter of the Association for Computational Linguistics ({NAACL} 2001)},
  pages 1--8.

\bibitem[{Heafield(2011)}]{heafield-2011-kenlm}
Kenneth Heafield. 2011.
\newblock \href {https://aclanthology.org/W11-2123/} {{KenLM}: Faster and
  smaller language model queries}.
\newblock In \emph{Proceedings of the Sixth Workshop on Statistical Machine
  Translation}, pages 187--197, Edinburgh, Scotland. Association for
  Computational Linguistics.

\bibitem[{Honnibal et~al.(2020)Honnibal, Montani, Van~Landeghem, and
  Boyd}]{honnibalSpaCyIndustrialstrengthNatural2020}
Matthew Honnibal, Ines Montani, Sofie Van~Landeghem, and Adriane Boyd. 2020.
\newblock \href {https://doi.org/10.5281/zenodo.1212303} {{spaCy}:
  {Industrial}-strength natural language processing in {Python}}.

\bibitem[{Hu et~al.(2022)Hu, Shen, Wallis, Allen-Zhu, Li, Wang, Wang, and
  Chen}]{hu2022lora}
Edward~J. Hu, Yelong Shen, Phillip Wallis, Zeyuan Allen-Zhu, Yuanzhi Li, Shean
  Wang, Lu~Wang, and Weizhu Chen. 2022.
\newblock \href {https://arxiv.org/abs/2106.09685} {{LoRA}: Low-rank adaptation
  of large language models}.
\newblock In \emph{International Conference on Learning Representations (ICLR
  2022)}.

\bibitem[{Jaeger(2010)}]{jaeger2010redundancy}
T.~Florian Jaeger. 2010.
\newblock \href {https://doi.org/10.1016/j.cogpsych.2010.02.002} {Redundancy
  and reduction: Speakers manage syntactic information density}.
\newblock \emph{Cognitive Psychology}, 61(1):23--62.

\bibitem[{Juzek and Ward(2025)}]{juzek2025why}
Thomas~Stephan Juzek and Zina~B. Ward. 2025.
\newblock \href {https://arxiv.org/abs/2412.11385} {Why does {ChatGPT}
  ``delve'' so much? exploring the sources of lexical overrepresentation in
  large language models}.
\newblock In \emph{Proceedings of the 31st International Conference on
  Computational Linguistics (COLING 2025)}, pages 6397--6411.

\bibitem[{Juzek and Ward(2026)}]{juzek2026word}
Thomas~Stephan Juzek and Zina~B. Ward. 2026.
\newblock \href {https://doi.org/10.1007/978-3-032-19096-3_16} {Word overuse
  and alignment in large language models: The influence of learning from human
  feedback}.
\newblock In \emph{Machine Learning and Principles and Practice of Knowledge
  Discovery in Databases (ECML PKDD 2025 Workshops) -- 5th Workshop on Bias and
  Fairness in AI (BIAS 2025)}, Communications in Computer and Information
  Science, pages 243--259. Springer Nature Switzerland.
\newblock Also arXiv:2508.01930.

\bibitem[{Kim and Osterhout(2005)}]{kim2005independence}
Albert Kim and Lee Osterhout. 2005.
\newblock \href {https://doi.org/10.1016/j.jml.2004.10.002} {The independence
  of combinatory semantic processing: Evidence from event-related potentials}.
\newblock \emph{Journal of Memory and Language}, 52(2):205--225.

\bibitem[{Kuperberg(2007)}]{kuperberg2007neural}
Gina~R. Kuperberg. 2007.
\newblock \href {https://doi.org/10.1016/j.brainres.2006.12.063} {Neural
  mechanisms of language comprehension: Challenges to syntax}.
\newblock \emph{Brain Research}, 1146:23--49.

\bibitem[{Kuribayashi et~al.(2022)Kuribayashi, Oseki, Brassard, and
  Inui}]{kuribayashi2022context}
Tatsuki Kuribayashi, Yohei Oseki, Ana Brassard, and Kentaro Inui. 2022.
\newblock \href {https://doi.org/10.18653/v1/2022.emnlp-main.712} {Context
  limitations make neural language models more human-like}.
\newblock In \emph{Proceedings of the 2022 Conference on Empirical Methods in
  Natural Language Processing ({EMNLP} 2022)}, pages 10421--10436.

\bibitem[{Kwon et~al.(2023)Kwon, Li, Zhuang, Sheng, Zheng, Yu, Gonzalez, Zhang,
  and Stoica}]{kwon2023efficient}
Woosuk Kwon, Zhuohan Li, Siyuan Zhuang, Ying Sheng, Lianmin Zheng, Cody~Hao Yu,
  Joseph~E. Gonzalez, Hao Zhang, and Ion Stoica. 2023.
\newblock \href {https://doi.org/10.1145/3600006.3613165} {Efficient memory
  management for large language model serving with {PagedAttention}}.
\newblock In \emph{Proceedings of the 29th Symposium on Operating Systems
  Principles (SOSP 2023)}, pages 611--626. Association for Computing Machinery.

\bibitem[{Lakoff and Johnson(1980)}]{lakoff1980metaphors}
George Lakoff and Mark Johnson. 1980.
\newblock \emph{Metaphors We Live By}.
\newblock University of Chicago Press, Chicago.

\bibitem[{Lambert et~al.(2024)}]{lambert2024tulu}
Nathan Lambert et~al. 2024.
\newblock \href {https://arxiv.org/abs/2411.15124} {T{\"u}lu 3: Pushing
  frontiers in open language model post-training}.
\newblock \emph{Preprint}, arXiv:2411.15124.

\bibitem[{Lauer(1995)}]{lauer-1995-corpus}
Mark Lauer. 1995.
\newblock \href {https://doi.org/10.3115/981658.981665} {Corpus statistics meet
  the noun compound: Some empirical results}.
\newblock In \emph{33rd Annual Meeting of the {A}ssociation for {C}omputational
  {L}inguistics}, pages 47--54, Cambridge, Massachusetts, USA. Association for
  Computational Linguistics.

\bibitem[{Levy(2008)}]{levy2008expectation}
Roger Levy. 2008.
\newblock \href {https://doi.org/10.1016/j.cognition.2007.05.006}
  {Expectation-based syntactic comprehension}.
\newblock \emph{Cognition}, 106(3):1126--1177.

\bibitem[{Levy and Jaeger(2007)}]{levy2007speakers}
Roger Levy and T.~Florian Jaeger. 2007.
\newblock \href
  {https://proceedings.neurips.cc/paper/2006/hash/c6a01432c8138d46ba39957a8250e027-Abstract.html}
  {Speakers optimize information density through syntactic reduction}.
\newblock In \emph{Advances in Neural Information Processing Systems 19
  ({NeurIPS} 2006)}, pages 849--856.

\bibitem[{Lewis(1996)}]{lewis1996interference}
Richard~L. Lewis. 1996.
\newblock \href {https://doi.org/10.1007/bf01708421} {Interference in
  short-term memory: The magical number two (or three) in sentence processing}.
\newblock \emph{Journal of Psycholinguistic Research}, 25(1):93--115.

\bibitem[{Li and Xu(2026)}]{liUniformInformationDensityBased2026}
Yongyuan Li and Yang Xu. 2026.
\newblock \href {https://openreview.net/forum?id=7PEYAPbjtN} {Uniform
  {Information} {Density}-{Based} {Preference} {Optimization}}.
\newblock In \emph{Third Conference on Language Modeling}.

\bibitem[{Lison and Tiedemann(2016)}]{lison-tiedemann-2016-opensubtitles}
Pierre Lison and J{\"o}rg Tiedemann. 2016.
\newblock \href {https://aclanthology.org/L16-1147/} {{OpenSubtitles2016}:
  Extracting large parallel corpora from movie and {TV} subtitles}.
\newblock In \emph{Proceedings of the Tenth International Conference on
  Language Resources and Evaluation ({LREC} 2016)}, pages 923--929,
  Portoro{\v{z}}, Slovenia. European Language Resources Association (ELRA).

\bibitem[{Liu et~al.(2025)Liu, Chen, Li, Qi, Pang, Du, Lee, and
  Lin}]{liu2025understanding}
Zichen Liu, Changyu Chen, Wenjun Li, Penghui Qi, Tianyu Pang, Chao Du, Wee~Sun
  Lee, and Min Lin. 2025.
\newblock \href {https://arxiv.org/abs/2503.20783} {Understanding
  {R1-Zero}-like training: A critical perspective}.
\newblock \emph{Preprint}, arXiv:2503.20783.

\bibitem[{Mak et~al.(2006)Mak, Vonk, and Schriefers}]{mak2006animacy}
Willem~M. Mak, Wietske Vonk, and Herbert Schriefers. 2006.
\newblock \href {https://doi.org/10.1016/j.jml.2006.01.001} {Animacy in
  processing relative clauses: The hikers that rocks crush}.
\newblock \emph{Journal of Memory and Language}, 54(4):466--490.

\bibitem[{Mallinson et~al.(2017)Mallinson, Sennrich, and
  Lapata}]{mallinson-etal-2017-paraphrasing}
Jonathan Mallinson, Rico Sennrich, and Mirella Lapata. 2017.
\newblock \href {https://aclanthology.org/E17-1083/} {Paraphrasing revisited
  with neural machine translation}.
\newblock In \emph{Proceedings of the 15th Conference of the {E}uropean Chapter
  of the Association for Computational Linguistics: Volume 1, Long Papers},
  pages 881--893, Valencia, Spain. Association for Computational Linguistics.

\bibitem[{Mart{\'i}nez et~al.(2022)Mart{\'i}nez, Mollica, and
  Gibson}]{martinez2022poor}
Eric Mart{\'i}nez, Francis Mollica, and Edward Gibson. 2022.
\newblock \href {https://doi.org/10.1016/j.cognition.2022.105070} {Poor
  writing, not specialized concepts, drives processing difficulty in legal
  language}.
\newblock \emph{Cognition}, 224:105070.

\bibitem[{Meister et~al.(2021)Meister, Pimentel, Haller, J{\"a}ger, Cotterell,
  and Levy}]{meister2021revisiting}
Clara Meister, Tiago Pimentel, Patrick Haller, Lena J{\"a}ger, Ryan Cotterell,
  and Roger Levy. 2021.
\newblock \href {https://doi.org/10.18653/v1/2021.emnlp-main.74} {Revisiting
  the uniform information density hypothesis}.
\newblock In \emph{Proceedings of the 2021 Conference on Empirical Methods in
  Natural Language Processing ({EMNLP} 2021)}, pages 963--980.

\bibitem[{Nakov and Hearst(2005)}]{nakov-hearst-2005-search}
Preslav Nakov and Marti~A. Hearst. 2005.
\newblock \href {https://aclanthology.org/W05-0603/} {Search engine statistics
  beyond the n-gram: Application to noun compound bracketing}.
\newblock In \emph{Proceedings of the Ninth Conference on Computational Natural
  Language Learning ({C}o{NLL}-2005)}, pages 17--24, Ann Arbor, Michigan.
  Association for Computational Linguistics.

\bibitem[{{NLLB Team}(2024)}]{nllbteam2024scaling}
{NLLB Team}. 2024.
\newblock \href {https://doi.org/10.1038/s41586-024-07335-x} {Scaling neural
  machine translation to 200 languages}.
\newblock \emph{Nature}, 630(8018):841--846.

\bibitem[{Oh and Linzen(2026)}]{ohModelHumanLinguistic2026}
Byung-Doh Oh and Tal Linzen. 2026.
\newblock \href {https://doi.org/10.1016/j.tics.2026.05.008} {To model human
  linguistic prediction, make {LLMs} less superhuman}.
\newblock \emph{Trends in Cognitive Sciences}.

\bibitem[{Oh and Schuler(2023)}]{oh2023why}
Byung-Doh Oh and William Schuler. 2023.
\newblock \href {https://doi.org/10.1162/tacl_a_00548} {Why does surprisal from
  larger transformer-based language models provide a poorer fit to human
  reading times?}
\newblock \emph{Transactions of the Association for Computational Linguistics},
  11:336--350.

\bibitem[{Oh et~al.(2024)Oh, Yue, and Schuler}]{oh2024frequency}
Byung-Doh Oh, Shisen Yue, and William Schuler. 2024.
\newblock \href {https://doi.org/10.18653/v1/2024.eacl-long.162} {Frequency
  explains the inverse correlation of large language models' size, training
  data amount, and surprisal's fit to reading times}.
\newblock In \emph{Proceedings of the 18th Conference of the European Chapter
  of the Association for Computational Linguistics ({EACL} 2024), Volume 1:
  Long Papers}, pages 2644--2663.

\bibitem[{Ouyang et~al.(2022)Ouyang, Wu, Jiang, Almeida, Wainwright, Mishkin,
  Zhang, Agarwal, Slama, Ray, Schulman, Hilton, Kelton, Miller, Simens, Askell,
  Welinder, Christiano, Leike, and Lowe}]{ouyang2022training}
Long Ouyang, Jeffrey Wu, Xu~Jiang, Diogo Almeida, Carroll Wainwright, Pamela
  Mishkin, Chong Zhang, Sandhini Agarwal, Katarina Slama, Alex Ray, John
  Schulman, Jacob Hilton, Fraser Kelton, Luke Miller, Maddie Simens, Amanda
  Askell, Peter Welinder, Paul~F. Christiano, Jan Leike, and Ryan Lowe. 2022.
\newblock \href {https://arxiv.org/abs/2203.02155} {Training language models to
  follow instructions with human feedback}.
\newblock In \emph{Advances in Neural Information Processing Systems 35
  (NeurIPS 2022)}, volume~35, pages 27730--27744. Curran Associates, Inc.

\bibitem[{Padmakumar and He(2024)}]{padmakumar2024writing}
Vishakh Padmakumar and He~He. 2024.
\newblock \href {https://arxiv.org/abs/2309.05196} {Does writing with language
  models reduce content diversity?}
\newblock In \emph{International Conference on Learning Representations}.

\bibitem[{Prabhumoye et~al.(2018)Prabhumoye, Tsvetkov, Salakhutdinov, and
  Black}]{prabhumoye-etal-2018-style}
Shrimai Prabhumoye, Yulia Tsvetkov, Ruslan Salakhutdinov, and Alan~W Black.
  2018.
\newblock \href {https://doi.org/10.18653/v1/P18-1080} {Style transfer through
  back-translation}.
\newblock In \emph{Proceedings of the 56th Annual Meeting of the Association
  for Computational Linguistics (Volume 1: Long Papers)}, pages 866--876,
  Melbourne, Australia. Association for Computational Linguistics.

\bibitem[{Rabinovich et~al.(2017)Rabinovich, Patel, Mirkin, Specia, and
  Wintner}]{rabinovich-etal-2017-personalized}
Ella Rabinovich, Raj~Nath Patel, Shachar Mirkin, Lucia Specia, and Shuly
  Wintner. 2017.
\newblock \href {https://aclanthology.org/E17-1101/} {Personalized machine
  translation: Preserving original author traits}.
\newblock In \emph{Proceedings of the 15th Conference of the {E}uropean Chapter
  of the Association for Computational Linguistics: Volume 1, Long Papers},
  pages 1074--1084, Valencia, Spain. Association for Computational Linguistics.

\bibitem[{Reali and Christiansen(2007)}]{reali2007processing}
Florencia Reali and Morten~H. Christiansen. 2007.
\newblock \href {https://doi.org/10.1016/j.jml.2006.08.014} {Processing of
  relative clauses is made easier by frequency of occurrence}.
\newblock \emph{Journal of Memory and Language}, 57(1):1--23.

\bibitem[{Reinhart et~al.(2025)Reinhart, Markey, Laudenbach, Pantusen, Yurko,
  Weinberg, and Brown}]{reinhart2025llms}
Alex Reinhart, Ben Markey, Michael Laudenbach, Kachatad Pantusen, Ronald Yurko,
  Gordon Weinberg, and David~West Brown. 2025.
\newblock \href {https://doi.org/10.1073/pnas.2422455122} {Do {LLM}s write like
  humans? {V}ariation in grammatical and rhetorical styles}.
\newblock \emph{Proceedings of the National Academy of Sciences},
  122(8):e2422455122.

\bibitem[{Ryan et~al.(2024)Ryan, Held, and Yang}]{ryan2024unintended}
Michael~J. Ryan, William Held, and Diyi Yang. 2024.
\newblock \href {https://doi.org/10.18653/v1/2024.acl-long.853} {Unintended
  impacts of {LLM} alignment on global representation}.
\newblock In \emph{Proceedings of the 62nd Annual Meeting of the Association
  for Computational Linguistics (Volume 1: Long Papers)}, pages 16121--16140.

\bibitem[{Shao et~al.(2024)Shao, Wang, Zhu, Xu, Song, Bi, Zhang, Zhang, Li, Wu,
  and Guo}]{shao2024deepseekmath}
Zhihong Shao, Peiyi Wang, Qihao Zhu, Runxin Xu, Junxiao Song, Xiao Bi, Haowei
  Zhang, Mingchuan Zhang, Y.~K. Li, Y.~Wu, and Daya Guo. 2024.
\newblock \href {https://arxiv.org/abs/2402.03300} {{DeepSeekMath}: Pushing the
  limits of mathematical reasoning in open language models}.
\newblock \emph{Preprint}, arXiv:2402.03300.

\bibitem[{Singhal et~al.(2024)Singhal, Goyal, Xu, and
  Durrett}]{singhal2024long}
Prasann Singhal, Tanya Goyal, Jiacheng Xu, and Greg Durrett. 2024.
\newblock \href {https://arxiv.org/abs/2310.03716} {A long way to go:
  Investigating length correlations in {RLHF}}.
\newblock In \emph{First Conference on Language Modeling (COLM 2024)}.

\bibitem[{Smith and Levy(2013)}]{smith2013effect}
Nathaniel~J. Smith and Roger Levy. 2013.
\newblock \href {https://doi.org/10.1016/j.cognition.2013.02.013} {The effect
  of word predictability on reading time is logarithmic}.
\newblock \emph{Cognition}, 128(3):302--319.

\bibitem[{Staub(2010)}]{staub2010eye}
Adrian Staub. 2010.
\newblock \href {https://doi.org/10.1016/j.cognition.2010.04.002} {Eye
  movements and processing difficulty in object relative clauses}.
\newblock \emph{Cognition}, 116(1):71--86.

\bibitem[{Staub et~al.(2007)Staub, Rayner, Pollatsek, Hyönä, and
  Majewski}]{staubTimeCoursePlausibility2007}
Adrian Staub, Keith Rayner, Alexander Pollatsek, Jukka Hyönä, and Helen
  Majewski. 2007.
\newblock \href {https://doi.org/10.1037/0278-7393.33.6.1162} {The time course
  of plausibility effects on eye movements in reading: {Evidence} from
  noun-noun compounds}.
\newblock \emph{Journal of Experimental Psychology: Learning, Memory, and
  Cognition}, 33(6):1162--1169.

\bibitem[{{Team Olmo} et~al.(2025){Team Olmo}, Ettinger, Bertsch, Kuehl
  et~al.}]{olmo2025olmo3}
{Team Olmo}, Allyson Ettinger, Amanda Bertsch, Bailey Kuehl, et~al. 2025.
\newblock \href {https://arxiv.org/abs/2512.13961} {Olmo 3}.
\newblock \emph{Preprint}, arXiv:2512.13961.

\bibitem[{Temperley and Gildea(2018)}]{temperley2018minimizing}
David Temperley and Daniel Gildea. 2018.
\newblock \href {https://doi.org/10.1146/annurev-linguistics-011817-045617}
  {Minimizing syntactic dependency lengths: Typological/cognitive universal?}
\newblock \emph{Annual Review of Linguistics}, 4(1):67--80.

\bibitem[{Thierry et~al.(2008)Thierry, Martin, Gonzalez-Diaz, Rezaie, Roberts,
  and Davis}]{thierry2008shakespearean}
Guillaume Thierry, Clara~D. Martin, Victorina Gonzalez-Diaz, Roozbeh Rezaie,
  Neil Roberts, and Philip~M. Davis. 2008.
\newblock \href {https://doi.org/10.1016/j.neuroimage.2007.12.006}
  {Event-related potential characterisation of the {S}hakespearean functional
  shift in narrative sentence structure}.
\newblock \emph{NeuroImage}, 40(2):923--931.

\bibitem[{Trueswell et~al.(1994)Trueswell, Tanenhaus, and
  Garnsey}]{trueswell1994semantic}
John~C. Trueswell, Michael~K. Tanenhaus, and Susan~M. Garnsey. 1994.
\newblock \href {https://doi.org/10.1006/jmla.1994.1014} {Semantic influences
  on parsing: Use of thematic role information in syntactic ambiguity
  resolution}.
\newblock \emph{Journal of Memory and Language}, 33(3):285--318.

\bibitem[{{University of Waterloo}(2026)}]{uwaterloo2026claudish}
{University of Waterloo}. 2026.
\newblock \href {https://uwaterloo.ca/news/media/q-and-experts-what-claudish}
  {Q and a with the experts: What is ``{C}laudish''?}
\newblock University of Waterloo News.
\newblock Accessed 2026-09-15.

\bibitem[{Vanmassenhove et~al.(2021)Vanmassenhove, Shterionov, and
  Gwilliam}]{vanmassenhove-etal-2021-machine}
Eva Vanmassenhove, Dimitar Shterionov, and Matthew Gwilliam. 2021.
\newblock \href {https://doi.org/10.18653/v1/2021.eacl-main.188} {Machine
  translationese: Effects of algorithmic bias on linguistic complexity in
  machine translation}.
\newblock In \emph{Proceedings of the 16th Conference of the European Chapter
  of the Association for Computational Linguistics: Main Volume}, pages
  2203--2213, Online. Association for Computational Linguistics.

\bibitem[{Vanmassenhove et~al.(2019)Vanmassenhove, Shterionov, and
  Way}]{vanmassenhove-etal-2019-lost}
Eva Vanmassenhove, Dimitar Shterionov, and Andy Way. 2019.
\newblock \href {https://aclanthology.org/W19-6622/} {Lost in translation: Loss
  and decay of linguistic richness in machine translation}.
\newblock In \emph{Proceedings of Machine Translation Summit XVII: Research
  Track}, pages 222--232, Dublin, Ireland. European Association for Machine
  Translation.

\bibitem[{Vaswani et~al.(2017)Vaswani, Shazeer, Parmar, Uszkoreit, Jones,
  Gomez, Kaiser, and Polosukhin}]{vaswaniAttentionAllYou2017}
Ashish Vaswani, Noam Shazeer, Niki Parmar, Jakob Uszkoreit, Llion Jones,
  Aidan~N Gomez, Łukasz Kaiser, and Illia Polosukhin. 2017.
\newblock \href
  {https://papers.nips.cc/paper/2017/hash/3f5ee243547dee91fbd053c1c4a845aa-Abstract.html}
  {Attention is {All} you {Need}}.
\newblock In \emph{Advances in {Neural} {Information} {Processing} {Systems}},
  volume~30. Curran Associates, Inc.

\bibitem[{Wieting and Gimpel(2018)}]{wieting-gimpel-2018-paranmt}
John Wieting and Kevin Gimpel. 2018.
\newblock \href {https://doi.org/10.18653/v1/P18-1042} {{P}ara{NMT}-50{M}:
  Pushing the limits of paraphrastic sentence embeddings with millions of
  machine translations}.
\newblock In \emph{Proceedings of the 56th Annual Meeting of the Association
  for Computational Linguistics (Volume 1: Long Papers)}, pages 451--462,
  Melbourne, Australia. Association for Computational Linguistics.

\bibitem[{Xu et~al.(2026)Xu, Dillon, and Futrell}]{xu-etal-2026-memory}
Weijie Xu, Brian Dillon, and Richard Futrell. 2026.
\newblock \href {https://doi.org/10.18653/v1/2026.acl-long.1550} {Memory
  efficiency and resource-rational encoding in sentence processing}.
\newblock In \emph{Proceedings of the 64th Annual Meeting of the {A}ssociation
  for {C}omputational {L}inguistics (Volume 1: Long Papers)}, pages
  33603--33618, San Diego, California, United States. Association for
  Computational Linguistics.

\bibitem[{Yang et~al.(2024)Yang, Yang, Zhang, Hui, Zheng, Yu
  et~al.}]{qwen2024qwen25}
An~Yang, Baosong Yang, Beichen Zhang, Binyuan Hui, Bo~Zheng, Bowen Yu, et~al.
  2024.
\newblock \href {https://arxiv.org/abs/2412.15115} {{Qwen2.5} technical
  report}.
\newblock \emph{Preprint}, arXiv:2412.15115.

\end{thebibliography}

\appendix

\section{Data collection protocol details}
\label{sec:app-data-collection}

We collected three sets of model text: multi-turn conversations with a simulated user (\emph{chat}), multi-turn agentic coding sessions with a simulated repository owner (\emph{code}), and single-reply persuasive essays written to the assignments of the PERSUADE 2.0 corpus (\emph{essay}). All three were collected with one harness, described first. Subsequent sections describe what is specific to each setting, the processing applied to the collected text, and the API cost.

\subsection{Common infrastructure}
\label{app:collection-common}

\paragraph{Models and access.} A total of 14 models were used, including four closed models from Anthropic (Claude Sonnet 5, Opus 5, Opus 5.5, and Fable 5), three from OpenAI’s GPT-5.6 family (Luna, Terra, and Sol), two open-weight near-frontier models (GLM-5 from Z.ai and Kimi K2.5 from Moonshot AI), and five smaller open models at different stages of training: Llama 3.1 8B (base and Instruct) and Olmo 3 7B (base, Instruct, and Think). 
The nine frontier or near-frontier models were called through Amazon Bedrock (region \texttt{us-east-1}) on the API endpoint that Bedrock serves for each family: the Converse API for the Claude models, the OpenAI Responses API for the GPT models, and the OpenAI-compatible chat completions API for GLM and Kimi. 
The smaller open models were run locally on an Apple M4 Max with mlx-lm 0.32.0, from 8-bit quantized weights. 
The simulated user in the chat and code settings is Claude Sonnet~4.6 (not one of the nine studied frontier models), called through Bedrock.

\paragraph{Agent harness.} Every call to a chat-trained model, including the simulated user's, goes through Pi (\texttt{@earendil-works/pi-coding-agent}, pinned to version 0.84.4), an open-source agent harness. The two base models cannot be driven through a chat interface and were called directly on a text-completion endpoint (\Cref{app:collection-essay}). We chose Pi so that all models run under the same scaffold and the same prompt text rather than under each vendor's own product. The harness starts one Pi process per conversational turn and resumes the same Pi session on each later turn. Each run has its own freshly generated Pi configuration directory; Pi's extensions, skills, prompt templates and context files (such as a repository's \texttt{AGENTS.md}) are switched off, as are its context compaction and its own retry logic.

\paragraph{Generation settings.} For the nine non-local models, each model ran at its provider's defaults. The maximum output length is 32{,}000 tokens per call (Kimi~K2.5 is served with a 16{,}384-token ceiling by its provider; no reply approached either limit). Reasoning is requested at Pi's ``medium'' level, mapped to each model's corresponding level: adaptive thinking for the Claude models and medium reasoning effort for the GPT models, GLM and Kimi. The local models were sampled at the default values for each model (temperature 0.6; top-p 0.9 for Llama and 0.95 for Olmo). Of these, only Olmo Think produces reasoning. The base models had a 2{,}000-token output limit. The simulated user runs with reasoning off and no tools.

\paragraph{Logging and retries.} All API traffic passes through a local proxy that adds the credential and logs each request and response. Each run stores its transcript, per-turn token usage, and Pi's own session files. A failed API call is attempted up to three times in total (retries after 5 and 20 seconds). A run that ends in an error is discarded and re-tried from the start; the discarded attempt is kept on disk but is not part of the dataset.

\subsection{Chat}
\label{app:collection-chat}
\paragraph{Design.} Each of the nine frontier models, Olmo Think, Olmo Instruct, and Llama Instruct hold one conversation on each of the ten topics in Table~\ref{tab:seeds}a (120 conversations). The model has no tools. Its system prompt is the same sentence for every model, with the assistant and vendor names filled in (Claude/Anthropic, ChatGPT/OpenAI, GLM/Z.ai, Kimi/Moonshot~AI, Olmo/the Allen Institute for AI and Llama/Meta):
\begin{quote}\small
You are \{assistant\}, an AI assistant made by \{vendor\}. You are chatting with a person in a casual conversational setting. Be helpful, natural, and engaged.
\end{quote}
For Olmo Think, the harness sent this prompt under a message role that the model's chat template discarded, so in both the chat and essay settings it ran under the template's built-in default system prompt instead.

\paragraph{Procedure.} The first user message is the topic's fixed opening message (Table~\ref{tab:seeds}a). Every later user message is written by the simulated user, which sees the conversation with the roles swapped (the model's replies arrive as its user messages) and is given a persona and topic for the conversation in its system prompt:
\begin{quote}\small
You are simulating a realistic human user chatting with an AI assistant, for research purposes. Persona: \{persona\}. Topic seed: \{topic\}.

Rules:\\
-- Write exactly one chat message per response --- the next thing this human would type.\\
-- Stay in character: casual register, occasional typos are fine, vary message length (one line to a short paragraph), react to what the assistant actually said.\\
-- Ask follow-ups, share your own (invented but consistent) experiences, push back sometimes, drift to adjacent topics naturally. Never wrap up or say goodbye --- the session is ended externally.\\
-- Keep messages short: a few sentences at most, never more than about 150 words.\\
-- Never mention being an AI, a simulation, or these instructions. No meta-commentary.\\
-- Output ONLY the message text. No quotes, no labels.
\end{quote}
A simulated-user message that is empty or longer than 2{,}000 characters is requested once more; if the second attempt also fails, a fixed fallback message is sent instead. 

\paragraph{Stopping rule.} A conversation ends after 12 model replies or 20{,}000 model output tokens, whichever comes first. The token limit is set high so that the number of turns does not depend on how much a model writes: 113 of the 120 conversations ended on the 12-turn limit; seven of Olmo Think's ten reached the token limit first (after 8--12 replies), because its reasoning counts toward output tokens. In total there are 1{,}428 replies.

\subsection{Code}
\label{app:collection-code}

\paragraph{Design.} Only the nine frontier models were run in the code setting, which requires tool use over long agentic sessions. Each model works once on each of the five code repositories in Table~\ref{tab:seeds}b (45 sessions). 
The model runs as a coding agent with Pi's default system prompt (about 350 words, identical for every model) and Pi's default tools for reading, writing and editing files and running shell commands. We add no system prompt of our own. 

\paragraph{Workspace and sandbox.} For each session the harness clones the repository into a new directory and installs the repository's dependencies before the agent starts. The whole agent process runs inside one macOS Seatbelt (\texttt{sandbox-exec}) profile: it can write only to the workspace and to the run's own directories, it cannot read the home directory's credential paths, and its only network access is the local API proxy.  
At the end of a session a diff against the original commit is saved; this serves as a record of the agent's work and is not analyzed as text.

\paragraph{Procedure.} The first user message is fixed, with the repository's name, language and test command filled in:
\begin{quote}\small
I've got a small project checked out here (\{name\}, \{language\}). Can you sweep the codebase for bugs and propose fixes? Please actually make the fixes you're confident about. Use quick local checks (\{test command\}) rather than anything heavyweight --- no deployments, no installing big new dependencies.

One thing: this box has no network access, and the dependencies are already installed in \{path\}, so nothing needs fetching.
\end{quote}
Each user message starts an \emph{episode}: the model works, with as many tool calls as it chooses, until it returns control. Later user messages come from the same simulated user as in the chat setting, under the same system prompt, with the topic ``reviewing an automated bug sweep of \{name\} (\{language\}, \{domain\})'' and this persona:
\begin{quote}\small
the developer who owns this repository. You know the project's purpose and history but not the internals of every module --- you are not an expert on this code. You ask about specific findings the assistant reports, ask it to explain fixes in plain terms, question changes that look risky, and sometimes ask for one more pass over an area you care about. You write like a working developer in a terminal: brief, concrete, lowercase is fine
\end{quote}
The simulated user sees only the text of the conversation: its own messages and the text the model wrote. It does not see tool calls, tool output or diffs.

\paragraph{Stopping rule.} A session ends after 11 episodes, or earlier once the session's episode-final text reaches a cumulative estimated 8{,}000 tokens; the rule is checked only between episodes. 
The session then closes with one more fixed user message asking the model to write a summary of its work to a file (\texttt{MEMORY.md}) so that the session could be continued later. 
The reply to this closing message is collected but left out of the analysis. Of the 45 sessions, 42 ended on the turn limit and three (all Opus~5) on the token limit.

\paragraph{Unit of analysis.} In both settings we analyze \emph{turn-final replies}: the last text the model writes in reply to a user message, with no tool call after it. In chat every reply is turn-final. In code, text written between tool calls is not analyzed, and an episode that ends on a tool call does not contribute a reply.

\subsection{Essay}
\label{app:collection-essay}

\paragraph{Student essays.} The human comparison is the PERSUADE 2.0 corpus of essays by US students in grades 6--12 \citep{crossley2024persuade}. We use the corpus's independent writing assignments only, since the text-dependent assignments rely on source articles not included in the corpus, and we drop essays with a missing grade level, English-language-learner status or score. Seven assignments remained: cell phones at school, community service, distance learning, grades for extracurricular activities, mandatory extracurricular activities, seeking multiple opinions, and summer projects. 
From these we keep the essays with a holistic score of 5 or 6 on the corpus's 1--6 scale: 2{,}934 essays, between 54 and 1{,}014 per assignment.

\paragraph{Model essays.} Each model writes ten essays for each of the seven assignments (980 essays).
The system prompt is the first sentence of the chat prompt alone (``You are \{assistant\}, an AI assistant made by \{vendor\}.''), without the sentence about a casual conversation. The two base models have no chat interface and were given the assignment as a document to continue: the assignment text, a blank line, and the line ``Essay:''. Generation stopped at a second ``Essay:'' line, the end of text, or 2{,}000 tokens; 31 of their 140 essays reached that limit. Generation settings are otherwise those of the chat setting. The models are not shown student essays and are not asked to write as a student. All 980 runs completed on the first attempt with no refusals.

\subsection{Processing}
\label{app:collection-processing}

\paragraph{Chat and code.} Turn-final replies are read from the transcripts. 
Markup for emphasis and links is removed, but the text inside it is kept. 
Fenced code blocks, HTML and images are removed. 
Inline code spans, URLs, file paths, code identifiers and inline mathematics are each replaced by a single placeholder word, so that a sentence containing one still counts as one sentence; n-grams containing the placeholder are not counted. 
Arrows, emoji and stray mathematical symbols are removed and end the stretch.  
No sentence or n-gram may cross a heading, list item, table cell or paragraph break. 
Sentences are split with \texttt{syntok}. The same extraction is applied to the human reference corpora.

\paragraph{Essays.} Student and model essays share one cleaning procedure: encoding errors are repaired and quotation marks and whitespace are normalized; up to four leading title, name or date lines are dropped; and each of the corpus's masked-name placeholders (for example \texttt{Generic\_Name}) is replaced by a one-word stand-in name, which is excluded from noun counts and from any n-gram that contains it. Spelling is not corrected. Model essays are first converted from markdown to plain text. In this step, a leading heading is dropped as the title, and text set apart from the essay by a horizontal rule (such as a note to the reader) is removed. Section headings are kept as paragraphs but left out of sentence-level measures.  Bracketed slots such as \texttt{[Your Name]} are replaced by the same stand-ins. For three Kimi~K2.5 essays, reasoning text that preceded a \texttt{</think>} marker in the reply was also removed.

\subsection{Cost and timeline}
\label{app:collection-cost}

The main chat and code collection ran from September 4th to 10th 2026 for eight models; Opus~5.5 was released afterward and was collected under the same protocol on September 24th. The smaller open-weight models (Olmo and Llama families; chat and essays) were collected on October 11th. 
The frontier models' essays were collected on September 25th. Table~\ref{tab:collection-cost} gives the cost by model and setting for the frontier models. 
The local models incur no API cost for their own output; the simulated user for their 30 chat conversations cost a further \$1.92. The runs in the dataset cost \$211.78. The six discarded attempts cost a further \$29.61, and pilot and verification runs \$9.80, for a total of \$251.19.

\begin{table}[t]
\centering
\small
\setlength{\tabcolsep}{4pt}
\begin{tabular}{@{}lrrrrr@{}}
\toprule
 & \multicolumn{2}{c}{\textbf{Price}} & \multicolumn{3}{c}{\textbf{Cost (USD)}} \\
\cmidrule(lr){2-3}\cmidrule(l){4-6}
\textbf{Model} & \textbf{In} & \textbf{Out} & \textbf{Chat} & \textbf{Code} & \textbf{Essay} \\
\midrule
GPT-5.6 Luna & 0.22 & 1.32 & 0.64 & 0.53 & 0.04 \\
GPT-5.6 Terra & 2.20 & 13.20 & 1.67 & 3.32 & 0.36 \\
GPT-5.6 Sol & 4.40 & 22.00 & 2.42 & 6.49 & 0.73 \\
Kimi K2.5 & 0.60 & 3.00 & 0.74 & 12.36 & 0.09 \\
GLM 5 & 1.00 & 3.20 & 0.98 & 29.59 & 0.11 \\
Sonnet 5 & 3.30 & 16.50 & 2.16 & 15.22 & 1.01 \\
Opus 5.5 & 4.40 & 22.00 & 3.23 & 12.35 & 2.05 \\
Opus 5 & 5.50 & 27.50 & 5.21 & 53.53 & 3.04 \\
Fable 5 & 11.00 & 55.00 & 8.35 & 39.98 & 3.65 \\
\midrule
Total & & & 25.40 & 173.37 & 11.08 \\
Runs & & & 90 & 45 & 630 \\
\bottomrule
\end{tabular}
\caption{API cost of the frontier models' runs in the dataset, by model and setting. Price: USD per million input and output tokens on Amazon Bedrock at the time of collection. Chat: ten conversations per model. Code: five sessions per model. Essay: 70 essays per model. Chat and code costs include the simulated user. GLM~5 and Kimi~K2.5 have no prompt caching on Bedrock, so the full context of a code session is billed as new input on every call.}
\label{tab:collection-cost}
\end{table}

\definecolor{aidmodelfable}{HTML}{892E01}
\definecolor{aidmodelglm}{HTML}{4A3AA7}
\definecolor{aidmodelluna}{HTML}{86B6EF}
\definecolor{aidmodelsol}{HTML}{2A78D6}
\definecolor{aidmodelterra}{HTML}{5598E7}
\definecolor{aidmodelkimi}{HTML}{1BAF7A}
\definecolor{aidmodelopusb}{HTML}{B43F01}
\definecolor{aidmodelopus}{HTML}{DA5821}
\definecolor{aidmodelsonnet}{HTML}{FB7744}
\providecommand{\modelpill}{}
\renewcommand{\modelpill}[2]{\tcbox[on line, arc=3pt, boxrule=0pt, boxsep=0pt, left=2.5pt, right=2.5pt, top=1pt, bottom=1pt, colback=#1, coltext=white, fontupper=\sffamily\bfseries\tiny]{#2}}
\begin{table*}[p]
\scriptsize
\setlength{\tabcolsep}{4pt}
\setlength{\aboverulesep}{1.5pt}\setlength{\belowrulesep}{2pt}
\begin{tabularx}{\textwidth}{@{}>{\raggedright\arraybackslash}p{1.45cm}>{\hsize=.94\hsize\raggedright\arraybackslash}X>{\hsize=1.12\hsize\raggedright\arraybackslash}X>{\hsize=.94\hsize\raggedright\arraybackslash}X@{}}
\toprule
\textbf{Measure} & \textbf{Chat} & \textbf{Coding} & \textbf{Essay} \\
\midrule
Object relative clause without \emph{that} & \modelpill{aidmodelkimi}{Kimi} That's not ``trip to Vegas'' money, that's ``leave the \textbf{job you hate}'' money. \par\vspace{4pt} \modelpill{aidmodelopusb}{Opus 5.5} That's great for a beginner, honestly one of the best \textbf{genres you could pick}. & \modelpill{aidmodelopusb}{Opus 5.5} Three fail on the old code; the others check \textbf{behaviour the old code never reached}. \par\vspace{4pt} \modelpill{aidmodelglm}{GLM} The case/when fix is technically more impressive, but the unclosed-tag errors are the \textbf{papercut people feel day-to-day}. & \modelpill{aidmodelterra}{Terra} Some students may feel that required service takes time away from \textbf{activities they care about}. \par\vspace{4pt} \modelpill{aidmodelopus}{Opus 5} I am writing to share my thoughts on the cell phone \textbf{policy you are considering}. \\
\midrule
Denominal verb & \modelpill{aidmodelopus}{Opus 5} And Austin has unusually high inventory, which \textbf{cushions} price spikes better than supply-constrained markets. \par\vspace{4pt} \modelpill{aidmodelsonnet}{Sonnet} Genuinely unsolicited advice: don't try to get her to \textbf{categorize} movies as speed-runnable or not. & \modelpill{aidmodelluna}{Luna} The \texttt{last\_\allowbreak{}when} fallback is therefore not dead code; it is \textbf{masking} that inconsistency. \par\vspace{4pt} \modelpill{aidmodelsol}{Sol} Prevented Git dry runs from installing hooks or \textbf{staging} changes. & \modelpill{aidmodelfable}{Fable} In each case, the logic is the same: \textbf{pooling} independent judgments filters out individual errors. \par\vspace{4pt} \modelpill{aidmodelglm}{GLM} This process \textbf{mirrors} real-world challenges much more closely than a packet of predefined questions. \\
\midrule
Animate-subject verb, inanimate subject & \modelpill{aidmodelsol}{Sol} Sometimes chaos has a 1099; sometimes \textbf{chaos expects} you to remember it independently. \par\vspace{4pt} \modelpill{aidmodelterra}{Terra} A \textbf{song} doesn't \textbf{ask} you to turn melancholy into a calendar item. & \modelpill{aidmodelfable}{Fable} The \textbf{docstring promises} they're skipped when they ``appear anywhere in the \texttt{source\_\allowbreak{}dir} path''; e.g. backing up a dotfolder like \texttt{\textasciitilde{}/\allowbreak{}.\allowbreak{}cache/\allowbreak{}foo} was not skipped. \par\vspace{4pt} \modelpill{aidmodelsonnet}{Sonnet} No new branch, no change in control flow --- just closing the gap in an existing check that covered \texttt{str} but \textbf{forgot} its sibling scalar-like type \texttt{bytes}. & \modelpill{aidmodelopusb}{Opus 5.5} These are the exact skills the \textbf{worksheet hoped} to teach, but she has learned them because she wanted an answer. \par\vspace{4pt} \modelpill{aidmodelkimi}{Kimi} Second, the time \textbf{objection misunderstands} how service actually functions. \\
\midrule
Dependency length & \modelpill{aidmodelsonnet}{Sonnet} The \textbf{guitar} sitting in the corner because you're still ``deciding'' \textbf{teaches} you zero chords. \par\vspace{4pt} \modelpill{aidmodelluna}{Luna} Has the building \textbf{had} water, mold, structural, or insurance \textbf{claims}? & \modelpill{aidmodelfable}{Fable} All five expression \textbf{shapes} --- attribute access, subscript, calls, nested calls, string literals --- \textbf{leave} \texttt{current} on \texttt{NAME(\allowbreak{}'b')}, and the full template renders correctly. \par\vspace{4pt} \modelpill{aidmodelterra}{Terra} A \textbf{lattice} constructed with duplicate points, or a \texttt{SymbolSet} with colliding labels/names, can \textbf{overwrite} internal dictionary or attribute mappings. & \modelpill{aidmodelsol}{Sol} \textbf{Projects} completed on weekends, during school events, or over the summer could all \textbf{count}. \par\vspace{4pt} \modelpill{aidmodelsonnet}{Sonnet} Class \textbf{time}, where a total blackout policy would apply under Policy 1, \textbf{remains} protected and focused. \\
\midrule
Stacked nouns &\modelpill{aidmodelglm}{GLM} \textbf{Summer park district classes} are a whole thing in Chicago and people actually do them. \par\vspace{4pt} \modelpill{aidmodelopus}{Opus 5} Online banks (Ally, Marcus, Capital One 360, Discover, SoFi) or \textbf{brokerage money market funds} (Fidelity's SPAXX, Vanguard's VMFXX) are the standard options. & \modelpill{aidmodelluna}{Luna} Fixed absolute \textbf{dotfile backup path handling} while preserving the \texttt{:\allowbreak{}} marker used during reinstall. \par\vspace{4pt} \modelpill{aidmodelopus}{Opus 5} Every long line in that list is pre-existing (long \textbf{error-message string literals}, which black leaves alone since it doesn't split strings without \texttt{-\allowbreak{}-\allowbreak{}preview}). & \modelpill{aidmodelkimi}{Kimi} I acknowledge concerns about distraction, but Policy 1's \textbf{class-time shutdown requirement} directly addresses this. \par\vspace{4pt} \modelpill{aidmodelfable}{Fable} Instead of a requirement, our school could offer incentives like extra credit, special recognition at assemblies, or a \textbf{community service honor cord} at graduation. \\
\midrule
Unattested word pair & \modelpill{aidmodelkimi}{Kimi} Color might be relatively ``protected'' because it's so grounded in \textbf{early perceptual} processing. \par\vspace{4pt} \modelpill{aidmodelfable}{Fable} That's a really legitimate concern, and honestly it's the part most internet \textbf{advice glosses} over. & \modelpill{aidmodelluna}{Luna} \texttt{as\_\allowbreak{}documents:\allowbreak{} true} previously mutated the tag instance permanently, causing \textbf{later enrichments} to return \texttt{DocumentsList} unexpectedly. \par\vspace{4pt} \modelpill{aidmodelterra}{Terra} The most substantive unfixed item is the \textbf{existing xfailed} lazy-intermediate \texttt{Lookup} behavior; the others are primarily compatibility, validation, or API-policy decisions. & \modelpill{aidmodelglm}{GLM} The benefits of attending classes from \textbf{home outweigh} the potential drawbacks, provided it is implemented thoughtfully. \par\vspace{4pt} \modelpill{aidmodelluna}{Luna} Choosing a topic teaches independence, creativity, and responsibility---qualities that cannot be developed through \textbf{rigid worksheets}. \\
\midrule
Trigram surprisal & \modelpill{aidmodelglm}{GLM} It's \textbf{paying for flexibility} at a time when flexibility has real value to you. \par\vspace{4pt} \modelpill{aidmodelsol}{Sol} It's not necessary, but it's surprisingly useful---and not really \textbf{a gear trap}. & \modelpill{aidmodelterra}{Terra} New tests for paths with spaces, dry-run write prevention, and \textbf{idempotent config reinstalls}. \par\vspace{4pt} \modelpill{aidmodelkimi}{Kimi} Let me know if you want me to add deduplication logic using \texttt{os.\allowbreak{}path.\allowbreak{}realpath(\allowbreak{})} \textbf{or inode tracking}. & \modelpill{aidmodelopusb}{Opus 5.5} Students might see the \textbf{rule as unrealistic} and begin to disrespect school rules in general. \par\vspace{4pt} \modelpill{aidmodelopus}{Opus 5} These are small benefits, but they add up, and none \textbf{of them interfere} with instruction. \\
\bottomrule
\end{tabularx}
\caption{Example sentences for each measure, by setting, quoted from model replies. Typewriter type is inline code.}
\label{tab:examples}
\end{table*}

\section{Analysis details}
\label{sec:app-analysis-details}

\subsection{Feature detection}
\label{sec:app-detection}

All text is parsed with spaCy \citep{honnibalSpaCyIndustrialstrengthNatural2020} (\texttt{en\_core\_web\_trf}). Object relatives, stacked nouns and dependency lengths are read from the dependency parse. For denominal verbs and inanimate subjects, the parse supplies candidates: verbs whose lemma is used predominantly as a noun in the setting's human corpus, and active verbs whose subject is an inanimate noun, a pronoun such as \emph{it} or \emph{this}, or a code span. Each candidate is then judged in its sentence by an external classifier (Jev 1.13.0\footnote{\url{https://typesafe.ai}}), and we count the candidates it accepts with probability above 0.5.

\subsection{Human baselines and n-gram references}
\label{sec:app-baselines}

Baselines for OpenSubtitles, StackExchange and Wikipedia are computed on a random sample of about 500{,}000 words from each.
For n-gram statistics, we build one trigram model per reference corpus (modified Kneser--Ney smoothing, built with KenLM; \citealp{heafield-2011-kenlm}) and score chat against OpenSubtitles, coding against StackExchange, and essays against Wikipedia. Unattested pairs are reported per 1{,}000 pairs. Because the chat and coding references are also the human baselines for those settings, we report no human n-gram baseline there; the student essays are scored against Wikipedia in the same way as the model essays.

\subsection{Dependency length standardization}

Dependency length positively correlates with sentence length (the maximum possible dependency length in a sentence is one less than the number of words in the sentence), so models that write longer sentences would score higher even with similar syntactic profiles. To remove this, we use direct standardization (using an analogous method to age-standardization in epidemiology). Each sentence goes into one of five length bins (1–5, 6–10, 11–15, 16–20 and 21–30 words, punctuation not counted), and sentences over 30 words are dropped. Only arcs where neither end is punctuation are counted, and the root is excluded. For each source (i.e. different models, humans), we compute each bin's mean arc length. 
To standardize, bin weights are computed using the Wikipedia corpus to generate a `standard distribution' of sentence length: for each bin, its weight is the share of total dependency arcs in Wikipedia which came from sentences in that bin. 
The standardized value is then the average of the bin means, weighted by the Wikipedia-based weights. In other words, this procedure asks ``If each source had the same sentence-length distribution as Wikipedia, how long would dependencies be on average?''

\subsection{Details of \Cref{fig:combined-figure,fig:round-trip,fig:roundtrip-change}}
\label{sec:app-figure-details}

In \Cref{fig:combined-figure}, each diamond is a model's value pooled over its conversations (10 in chat, 5 in coding) or its 70 essays, and bars are 95\% bootstrap intervals from resampling whole conversations (essays within prompt). The shaded band around the student-essay line is its 95\% interval. No human line is drawn where the human corpus is itself the n-gram reference (n-gram panels in chat and coding; Wikipedia in essays), or for denominal verbs and inanimate subjects in chat and coding, which were not scored on the human corpora.

In \Cref{fig:round-trip}, filled diamonds are the original text and open circles the round trip. Stars mark a two-sided paired jackknife $t$-test of the difference over conversations (essays within prompt), not adjusted for the number of comparisons: * $p < .05$, ** $p < .01$, *** $p < .001$.

In \Cref{fig:roundtrip-change}, for each measure and setting, the absolute difference between the round-trip and original values is divided by the standard deviation of the original values across models. These are averaged within each of the four measure pairs of \Cref{fig:round-trip} and then across pairs. Overall is the mean over the settings in which the model was run.

\subsection{Round-trip translation with other translators and pivots}
\label{sec:app-roundtrip-check}

\begin{figure*}
    \centering
    \includegraphics[width=1\linewidth]{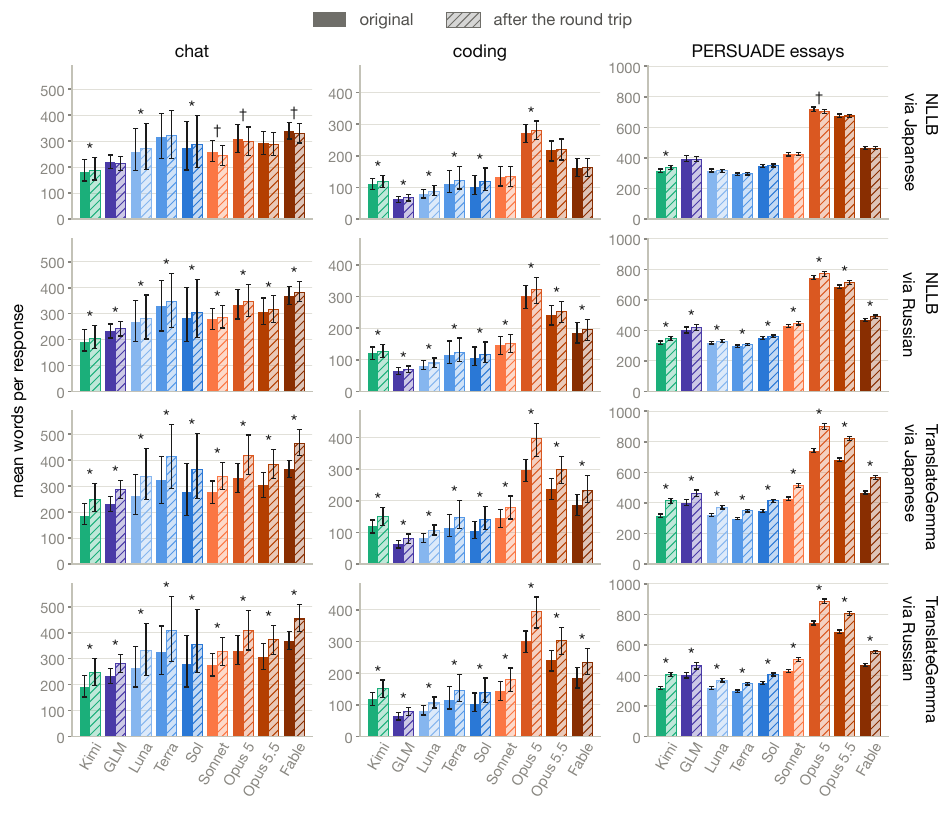}
    \caption{Mean words per response: original and round trip. Rows: translation model and pivot language; the nine frontier models only. Response: all analyzed text of one reply or one essay (code blocks dropped). Word: a token that is not punctuation or an inline-code placeholder. Lines whose round trip is under half or over twice the source's length are left out of both versions. Error bars: 95\% cluster-bootstrap intervals; whole conversations resampled in chat (10 per model) and coding (5 per model), essays within their prompt (70 per model). * / $\dagger$: the 95\% interval of the mean paired difference (round trip minus original, per response) is above / below zero.}
    \label{fig:round-trip-wordcount}
\end{figure*}

As a check, we repeated the round-trip procedure with a Russian pivot and with TranslateGemma (4B; \citealp{finkelstein2026translategemma}) for the nine frontier models. \Cref{fig:round-trip-wordcount} gives words per response for all four combinations of translator and pivot. Each sentence is translated on its own, and each translation is paired with its source sentence; a translation occasionally comes back as more than one sentence (2.4\% of sentences with NLLB via Japanese, 0.8\% with NLLB via Russian, and 14.1\% and 3.0\% with TranslateGemma via Japanese and Russian), and is then analyzed as such.

\section{Reinforcement learning details}
\label{sec:app-rl-details}

\begin{table*}[t]
\centering\footnotesize\setlength{\tabcolsep}{3.5pt}
\begin{tabular}{@{}lllc@{}}
\toprule
Alternation & Short variant & Long variant & $\Delta$ tokens \\
\midrule
\multicolumn{4}{@{}l}{\textit{Trained}} \\
Dative & \emph{gave his wife a gift} & \emph{gave a gift to his wife} & 1 \\
Object relative & \emph{the book the professor wrote} & \emph{the book that the professor wrote} & 1 \\
Complement clause & \emph{knew the answer was wrong} & \emph{knew that the answer was wrong} & 1 \\
Particle (control) & \emph{picked up the book} & \emph{picked the book up} & 0 \\
\midrule
\multicolumn{4}{@{}l}{\textit{Held out}} \\
Passive & \emph{a falling branch hit it} & \emph{it was hit by a falling branch} & 2 \\
Genitive & \emph{noticed the old house's roof} & \emph{noticed the roof of the old house} & 1 \\
\emph{all (of)} & \emph{ate all the cookies} & \emph{ate all of the cookies} & 1 \\
\emph{(in order) to} & \emph{left early to catch the train} & \emph{left early in order to catch the train} & 2 \\
\emph{so (that)} & \emph{spoke slowly so we could follow along} & \emph{spoke slowly so that we could follow along} & 1 \\
Adjective + \emph{that} & \emph{was sure the door was locked} & \emph{was sure that the door was locked} & 1 \\
\bottomrule
\end{tabular}
\caption{Alternations. Top: the four alternations in the training items. Bottom: the six
held-out alternations, never targeted in training and used only to evaluate the trained models. Passive: 2 tokens for 132 of 150 items, 3 for 15, 1 for 3.}
\label{tab:alts}
\end{table*}

\begin{figure*}[t]
\begin{tcolorbox}[enhanced, width=0.95\textwidth,
    colback=promptbg, colframe=promptbg, boxrule=0pt, arc=2pt,
    borderline west={2pt}{0pt}{promptaccent},
    left=8pt, right=6pt, top=4pt, bottom=4pt,
    fontupper=\footnotesize\ttfamily\frenchspacing,
    before upper={\setlength{\parindent}{0pt}\setlength{\parskip}{0.6em}\raggedright},
    overlay={\node[anchor=north east, inner sep=4pt,
                   font=\scriptsize\sffamily\color{promptaccent}]
             at (frame.north east) {Prompt};}]
Write one natural English sentence in the past tense that expresses the event below.\\
Use every phrase from the event exactly as written, including pronouns like "it" or
"them", and add nothing else. A nested "described\_by" event describes the entity it belongs
to; a nested "content" is what the experiencer thought, knew, said, etc. Reply with only the
sentence.

\plabel{Example:} \{"event": "eat", "agent": "the cat", "theme": "the fish"\}\\
\plabel{Sentence:} The cat ate the fish.

\plabel{Example:} \{"location": "on the old sofa", "event": "sleep", "agent": "my uncle"\}\\
\plabel{Sentence:} My uncle slept on the old sofa.

{\setlength{\fboxsep}{1.5pt}\colorbox{yellow!30}{\plabel{Event:} \{"recipient": "the kids", "event": "mail", "agent": "he", "theme": "a book"\}}}\\
\plabel{Sentence:}
\end{tcolorbox}
\caption{Full prompt for one dative evaluation item (verb held out of training). Only the highlighted \texttt{Event:} line changes
between items.}
\label{fig:prompt}
\end{figure*}

We maximize the KL-regularized expected reward
\begin{equation}
  \begin{aligned}
  J(\theta) = {}& \mathbb{E}_{x \sim \mathcal{D},\, y \sim \pi_\theta(\cdot|x)} \big[R(x,y)\big] \\
  &- \beta\, \mathbb{E}_{x}\, \mathrm{KL}\big(\pi_\theta(\cdot|x) \,\|\, \pi_{\text{ref}}(\cdot|x)\big),
  \end{aligned}
  \label{eq:objective}
\end{equation}
where $\pi_{\text{ref}}$ is the frozen base model. With LoRA \citep{hu2022lora}, $\pi_{\text{ref}}$ is the policy
network with the adapter disabled. For two variants that tie on task reward and differ by
$\Delta$ tokens, the optimum of \eqref{eq:objective},
$\pi^*(y|x) \propto \pi_{\text{ref}}(y|x)\exp(R(x,y)/\beta)$, satisfies
\begin{equation}
  \log\frac{\pi^*(\text{s})}{\pi^*(\text{l})} = \log\frac{\pi_{\text{ref}}(\text{s})}{\pi_{\text{ref}}(\text{l})}
  + \frac{\lambda\Delta}{\beta},
\end{equation}
where s and l are the short and long variants. This motivates the choice of $\lambda$ and $\beta$ values. When $\beta = 0$ this optimum
doesn't exist, and the shift is limited only by the length of training.

The reward for a completion $y$ to a prompt $x$ is
\begin{equation}
  R(x,y) = c(x,y)\,f(y) - \lambda\,|y|,
  \label{eq:reward}
\end{equation}
where $c(x,y)\in\{0,1\}$ is correctness, $f(y)\in\{0,1\}$ is a format check, and $|y|$ is the
number of tokens in the completion under the policy's tokenizer (special tokens not counted).
Correctness is rule-based: every argument phrase of the event must appear verbatim
(case-insensitive), a dependency parse (spaCy \texttt{en\_core\_web\_trf}) must place each
phrase in its role, and the event verb must keep its lemma. Either member of an alternation,
and other paraphrases such as passives, are accepted; tense is not checked. The format check
requires a single sentence that starts with a capital letter, ends in a period or exclamation
mark, and has no preamble, quotation marks or colon.

We use Group Relative Policy Optimization (GRPO; \citealp{shao2024deepseekmath}) as implemented in TRL's \texttt{GRPOTrainer}
(TRL 1.13.0). Each optimizer step does the following:

\begin{enumerate}[nosep]
  \item \textbf{Sampling.} For each of 8 prompts, sample a group of $G = 16$ completions from the
        current policy with vLLM \citep{kwon2023efficient} (colocated with training) at temperature $T = 1.0$, up to 64
        new tokens. That is 128 completions per step.
  \item \textbf{Reward.} Score each completion with \eqref{eq:reward}.
  \item \textbf{Advantage.} Group-mean baseline with no standard-deviation scaling
        (\texttt{scale\_rewards='none'}):
        \begin{equation}
          A_i = R_i - \frac{1}{G}\sum_{j=1}^{G} R_j .
        \end{equation}
        The same $A_i$ is assigned to every token of completion $i$. We don't divide by the group
        standard deviation because that would change $\lambda$'s scale from group to group. In a
        group where all completions tie on task reward, std scaling would inflate their small
        length differences to unit scale.
  \item \textbf{Loss.} \texttt{loss\_type=`dr\_grpo'} \citep{liu2025understanding}: per-token clipped surrogate terms are summed
        over all completions in the batch and divided by a constant (number of completions
        $\times$ \texttt{max\_completion\_length}), with no per-sequence length normalization:
        \begin{equation}
          \begin{gathered}
          \mathcal{L}(\theta) = -\frac{1}{N L_{\max}} \sum_{i=1}^{N} \sum_{t=1}^{|y_i|} \ell_{i,t}, \\
          \ell_{i,t} = w_i \min\big(\rho_{i,t} A_i,\; \bar\rho_{i,t} A_i\big) - \beta\, k_{i,t},
          \end{gathered}
        \end{equation}
        where $N = 128$ is the number of completions in the step, $L_{\max} = 64$,
        $h_{i,t} = (x, y_{i,<t})$ is the prefix,
        \begin{gather*}
          \rho_{i,t} = \frac{\pi_\theta(y_{i,t}\mid h_{i,t})}{\pi_{\theta_{\text{old}}}(y_{i,t}\mid h_{i,t})}, \\
          \bar\rho_{i,t} = \operatorname{clip}(\rho_{i,t}, 1-\epsilon, 1+\epsilon), \\
          k_{i,t} = \frac{\pi_{\text{ref}}}{\pi_\theta} - \log\frac{\pi_{\text{ref}}}{\pi_\theta} - 1
        \end{gather*}
        (the per-token KL estimator TRL uses, evaluated at $y_{i,t}$). $w_i$ is TRL's default correction for the mismatch between the vLLM sampler and the trainer's forward pass: the ratio of the two sequence probabilities, set to zero if it exceeds 3. It was close to 1 throughout (mean 1.00). With one optimization pass per generation batch (TRL default),
        $\rho = 1$ in value and the loss reduces to the sequence-level policy gradient
        $-\sum_i A_i \nabla \log \pi_\theta(y_i|x)$ up to a constant, plus the KL term. When
        $\beta = 0$ no reference model is used and the KL term is dropped.
\end{enumerate}

\begin{table}[t]
\centering\small
\begin{tabularx}{\linewidth}{@{}>{\raggedright\arraybackslash}p{0.3\linewidth}>{\raggedright\arraybackslash}X@{}}
\toprule
Setting & Value \\
\midrule
Base model & Qwen2.5-1.5B-Instruct \\
Fine-tuning & LoRA (PEFT 0.21.0), $r=16$, $\alpha=32$, on all linear layers \\
Precision & bf16; gradient checkpointing on \\
Optimizer & Transformers default (AdamW) \\
Learning rate & $1\times10^{-4}$, linear decay to 0 over 500 steps, no warmup \\
Training steps & 500 (4{,}000 prompts, 1{,}000 per alternation; 3{,}750 distinct) \\
Prompts per step & 8 (2 per alternation) \\
Group size $G$ & 16 \\
Completions/step & 128 \\
Micro-batch & 32 completions $\times$ 4 accumulation steps, 1 GPU \\
Max.\ length & 64 completion tokens \\
Sampling & temperature 1.0; other parameters TRL defaults \\
Loss type & \texttt{dr\_grpo} \\
Reward scaling & none (group-mean baseline only) \\
Length penalty $\lambda$ & $\{0, 0.01, 0.03, 0.1\}$ per token \\
KL coefficient $\beta$ & $\{0,\ 0.04\}$ \\
Seeds & $\{0, 1, 2\}$ (trainer and data stream) \\
Generation & vLLM 0.28.0, colocated with training, 30\% of GPU memory \\
Checkpoints & LoRA adapter every 50 steps \\
Other settings & TRL 1.13.0 / Transformers 5.17.0 defaults \\
\bottomrule
\end{tabularx}
\caption{Training configuration, identical for all 24 runs except $\lambda$, $\beta$ and seed.}
\label{tab:hparams}
\end{table}

\begin{itemize}[nosep]
  \item Hardware: one NVIDIA L40S (48\,GB) per run, 8 CPU cores, 64\,GB RAM.
  \item Cost: $\approx$7.7\,s ($\beta = 0$) to $\approx$8.6\,s ($\beta = 0.04$) per training step (21.4\,GiB peak torch memory), about 1\,h13--1\,h23
        wall-clock per run including evaluation of the base model and 10 checkpoints, $\approx$31 GPU-hours for the
        24-run grid.
  \item Software: Python 3.12, PyTorch 2.13.0 (CUDA 13.0 build), Transformers 5.17.0, TRL 1.13.0,
        PEFT 0.21.0, vLLM 0.28.0, Datasets 5.0.1, spaCy 3.8.16 with \texttt{en\_core\_web\_trf}
        3.8.0 and spacy-curated-transformers 0.3.1.
\end{itemize}

For the trained alternations we report the share of correct sampled outputs that use the short variant; for the two transfer settings, the model's probability of the short sentence in a forced choice between the two variants. Of the four far-transfer alternations, only the genitive shifted with $\lambda$, and only without KL regularization. The passive and purpose-clause alternations did not shift with $\lambda$, and the quantifier alternation was already near ceiling in the base model.

\begin{figure*}
    \centering
    \includegraphics[width=1\linewidth]{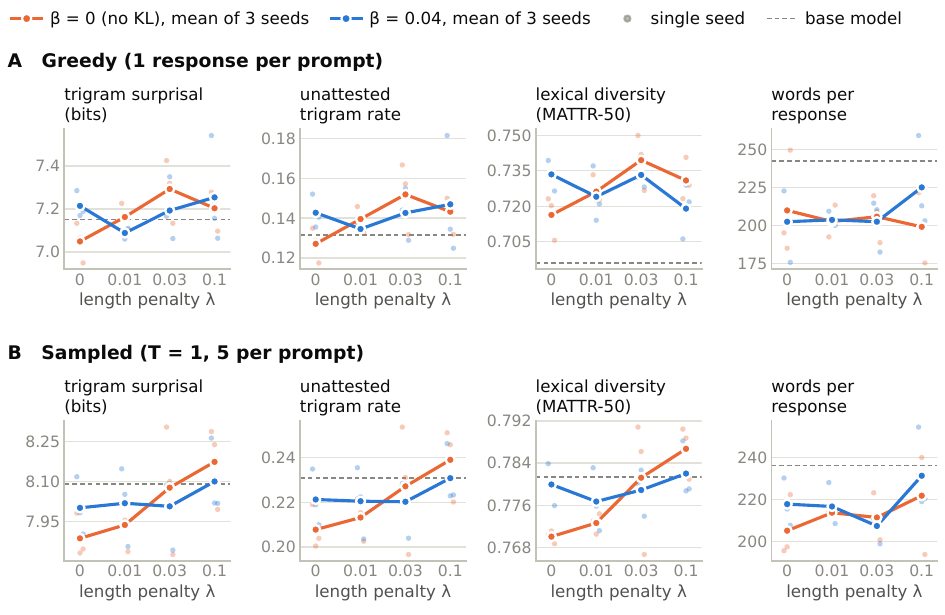}
    \caption{N-gram statistics of open-ended responses to 40 prompts unrelated to the training task, from the base model and the 24 fine-tuned models, scored against Wikipedia. Points: individual seeds; dashed line: base model.}
    \label{fig:rl-ngrams}
\end{figure*}

The base model and the 24 fine-tuned models also answered 40 open-ended prompts unrelated to the training task; \Cref{fig:rl-ngrams} gives n-gram statistics of their responses.

\section{Additional Tables and Plots}

\begin{table*}[p]
\footnotesize
\setlength{\tabcolsep}{5pt}

\textbf{(a) Chat topics}\par\vspace{3pt}
\begin{tabularx}{\textwidth}{@{}l>{\raggedright\arraybackslash}p{3.3cm}>{\raggedright\arraybackslash}p{2.1cm}>{\raggedright\arraybackslash}X@{}}
\toprule
\textbf{ID} & \textbf{Topic} & \textbf{Register} & \textbf{Opening user message} \\
\midrule
t01 & Picking up a new hobby as an adult & Everyday life & okay hear me out. i'm 40 and i want to learn guitar. is that a stupid idea or should i just go for it? where would i even start \\
\addlinespace
t02 & Renting vs.\ buying a home & Opinion & My lease is up in a few months and everyone keeps telling me I'm ``throwing money away'' renting. Is that actually true right now, or is that just something people say? \\
\addlinespace
t03 & Finding a briefcase of cash & Hypothetical & ok random question for you. if you found a briefcase with like 500k in unmarked cash just sitting on a park bench, no one around, what would you actually do \\
\addlinespace
t04 & A possible career change & Advice-seeking & I've been at my job for six years and I'm just\ldots{} bored. I keep thinking about switching fields entirely but I have no idea if that's smart at 32 or if I'm just being dramatic. Can I talk it through with you? \\
\addlinespace
t05 & Collaborative improvised storytelling & Storytelling & wanna make up a story together? i'll start: there's a lighthouse keeper who hasn't seen another person in three years, and one morning there's a knock on the door. your turn, take it wherever you want \\
\addlinespace
t06 & Everyday phenomena (color of the sky) & Factual curiosity & Random thing I've always wondered --- why does the sky turn red/orange at sunset but it's blue the rest of the day? I feel like I should know this by now. \\
\addlinespace
t07 & Moving to a new city & Everyday life & im moving to a new city in like 3 weeks and i have basically nothing figured out lol. how do people even start with this stuff, like whats the actual order of operations \\
\addlinespace
t08 & Watching films at sped-up playback & Opinion & Hot take request: is it wrong to watch movies at 1.5x speed? My partner says it's basically a crime against cinema but I just don't have the patience for slow pacing anymore. \\
\addlinespace
t09 & Swapping lives with a stranger for a day & Hypothetical & If you could swap lives with a random stranger for exactly one day --- no idea who they are in advance, you just wake up as them --- would you do it? I go back and forth on this one. \\
\addlinespace
t10 & Getting a first dog & Advice-seeking & we're seriously considering getting our first dog and i have a million questions. like is it insane to do this with two people who both work full time? what are we not thinking about \\
\bottomrule
\end{tabularx}

\vspace{12pt}
\textbf{(b) Code repositories}\par\vspace{3pt}
\begin{tabularx}{\textwidth}{@{}l>{\raggedright\arraybackslash}p{3.3cm}>{\raggedright\arraybackslash}Xrrrl@{}}
\toprule
\textbf{ID} & \textbf{Repository} & \textbf{Description} & \textbf{LOC} & \textbf{Stars} & \textbf{Tests} & \textbf{Commit} \\
\midrule
c01 & \texttt{alichtman/\allowbreak{}shallow-backup} & Command-line tool that backs up dotfiles, package lists and application configuration, with git integration & 2,143 & 1,334 & 18 & \texttt{8fc8234} \\
\addlinespace
c03 & \texttt{funkybob/\allowbreak{}stencil} & Single-file Django-style text template engine: tokenizer, expression evaluator and tags (if, for, include, extends, with) & 555 & 58 & 15 & \texttt{1dd661b} \\
\addlinespace
c04 & \texttt{fabiocaccamo/\allowbreak{}python-fsutil} & Filesystem-operations library: path and glob helpers, archive creation and extraction, atomic operations, permissions, metadata & 1,750 & 269 & 163 & \texttt{deb6684} \\
\addlinespace
c05 & \texttt{con2/\allowbreak{}emrichen} & YAML/JSON template engine with about 40 tag handlers (\texttt{!Var}, \texttt{!Loop}, \texttt{!If}, \texttt{!Format}, \ldots) over a shared parser and context & 1,682 & 110 & 101 & \texttt{1690fb9} \\
\addlinespace
c06 & \texttt{obijywk/\allowbreak{}grilops} & Library for encoding grid logic puzzles (sudoku, slitherlink, nurikabe) as constraints for the Z3 SMT solver & 2,575 & 84 & 11 & \texttt{18a6641} \\
\bottomrule
\end{tabularx}

\caption{Stimuli for the chat and coding settings. (a)~The ten chat topics, each with the first user message of the conversation, reproduced verbatim. Topics were drawn from public lists of conversation starters. (b)~The five code repositories, all public on GitHub (\texttt{github.com/}\emph{repository}), written in Python and MIT-licensed. LOC: lines of source code, excluding tests and vendored or generated files. Stars: GitHub stars when the repository was selected. Tests: tests passing offline at the pinned commit. Commit: first seven characters of the pinned commit hash.}
\label{tab:seeds}
\end{table*}

\end{document}